\pdfoutput=1
\documentclass{article}

\usepackage[preprint,nonatbib]{nips_2018}

\usepackage[utf8]{inputenc} 
\usepackage[T1]{fontenc}    
\usepackage{url}            
\usepackage{booktabs}       
\usepackage{amsfonts}       
\usepackage{nicefrac}       
\usepackage{microtype}      

\def\psfancypar#1#2{\begingroup\def\par{\endgraf\endgroup\lineskiplimit=0pt}
               \setbox2=\hbox{\large\sc #2}
               \newdimen\tmpht \tmpht \ht2 \advance\tmpht by \baselineskip
               \font\hhuge=Times-Bold at \tmpht
               \setbox1=\hbox{{\hhuge #1}}
               \count7=\tmpht \count8=\ht1
               \divide\count8 by 1000 \divide\count7 by \count8 
               \tmpht=.001\tmpht\multiply\tmpht by \count7 
               \font\hhuge=Times-Bold at \tmpht
               \setbox1=\hbox{{\hhuge #1}}
               \noindent
                \hangindent1.05\wd1
               \hangafter=-2 {\hskip-\hangindent
               \lower1\ht1\hbox{\raise1.0\ht2\copy1}%
                \kern-0\wd1}\copy2\lineskiplimit=-1000pt}

\def\boxit#1{\vbox{\hrule\hbox{\vrule\kern3pt
        \vbox{\kern3pt#1\kern3pt}\kern3pt\vrule}\hrule}}

\def\reals{ { {\rm  I \kern-0.15em R }  } }
\def\complex{ {\,{{\rm C} \kern-0.50em \raise0.20ex {  |}}\, }}

\def\Rbf{{\bf R}}

\def\Ac{{\cal A}}

\def\Nc{{\cal N}}

\def\Sc{{\cal S}}

\def\be{\vskip .3cm \begin{equation}}
\def\ee{\end{equation} \vskip .4cm \noindent}
\def\Rxx{\Rbf_{\ssstyle X\kern-.1em X}}

\let\ssstyle=\scriptscriptstyle

\def\Kout{\setbox1=\hbox{\Huge\bf K}\hbox to
1.05\wd1{\hspace{.05\wd1}
}}
\def\Sout{\setbox1=\hbox{\Huge\bf S}\hbox to 1.05\wd1{\hspace{.05\wd1}
}}

\usepackage{cite}
\usepackage{amsmath,amssymb,amsfonts,latexsym,verbatim,color,epsfig,psfrag}
\usepackage{times,multirow,multicol, array}
\usepackage{algorithm,algorithmic}
\usepackage{setspace}
\usepackage{float}
\newcolumntype{C}[1]{>{\centering\let\newline\\\arraybackslash\hspace{0pt}}m{#1}}

\title{AMBER: Adaptive Multi-Batch Experience Replay for Continuous Action Control}
\author{
	Seungyul Han \\
	Dept. of Electrical Engineering\\
	KAIST\\
	Daejeon, South Korea 34141\\
	\texttt{sy.han@kaist.ac.kr} \\
	\And
	Youngchul Sung$^{\dagger}$ \\
	Dept. of Electrical Engineering\\
	KAIST \\
	Daejeon, South Korea 34141\\
	\texttt{ycsung@kaist.ac.kr} \\
}

\begin{document}

\maketitle

\begin{abstract}
In this paper,  a new adaptive  multi-batch experience replay scheme is proposed  for proximal policy optimization (PPO) for continuous action control.
On the contrary to original PPO, the proposed scheme uses the batch samples of  past policies as well as the current policy for the update for the next policy, where the number of the used past batches is adaptively determined based on the oldness of the past batches measured by the average importance sampling (IS) weight. The new algorithm constructed by combining PPO with the proposed multi-batch experience replay scheme  maintains the advantages of original PPO such as random mini-batch sampling and small bias due to low IS weights by storing the pre-computed advantages and values and adaptively determining the mini-batch size. Numerical results show that the proposed method significantly increases the speed and  stability of convergence on various continuous control tasks  compared to original PPO.
\end{abstract}

\section{Introduction} \label{sec:Introduction}

Reinforcement learning (RL) aims to optimize the policy for the cumulative reward in a Markov decision process (MDP) environment. SARSA and Q-learning are well-known RL algorithms for learning finite MDP environments, which store all Q values as a table and solve the Bellman equation \cite{watkins1992q,rummery1994line,sutton1998reinforcement}. However, if the state space of environment is infinite, all Q values cannot be stored. Deep Q-learning (DQN) \cite{mnih2013playing} solves this problem by using a Q-value neural network to approximate  and generalize  Q-values from finite experiences, and DQN is shown to outperform the human level in Atari games with discrete action spaces \cite{mnih2015human}.
For discrete action spaces, the policy simply can choose the optimal action that has the maximum Q-value, but this is not possible for continuous action spaces. Thus, policy gradient (PG) methods that parameterize the policy by using a neural network and optimize the parameterized policy to choose optimal action from the given Q-value are considered for continuous action control \cite{sutton2000policy}.
Recent PG methods can be classified mainly into two groups: 1) Value-based PG methods that update the policy to choose action by following the maximum distribution  or the exponential distribution of Q-value, e.g., deep deterministic policy gradient (DDPG) \cite{lillicrap2015continuous}, twin-delayed DDPG (TD3) \cite{fujimoto2018addressing}, and soft-actor critic (SAC) \cite{haarnoja2018soft}, and 2) IS-based PG methods that directly update the policy to maximize the discounted reward sum by using IS, e.g., trust region policy optimization (TRPO) \cite{schulman2015trust}, actor-critic with experience replay (ACER) \cite{wang2016sample}, PPO \cite{schulman2017proximal}. Both PG methods update the policy parameter by using stochastic gradient descent (SGD), but the convergence speed of SGD is slow since the gradient direction of SGD is unstable. Hence, increasing sample efficiency is important to PG methods for fast convergence.
Experience replay (ER), which was first considered in DQN \cite{mnih2013playing}, increases sample efficiency by storing old sample from the previous policies and reusing these old samples for current update, and enhances the learning stability by reducing the sample correlation by sampling random mini-batches from a large replay memory. For value-based PG methods, ER  can be applied without any modification, so state-of-the-art value-based algorithms (TD3, SAC) use ER. However, applying ER to IS-based PG methods is a challenging problem. For IS-based PG methods,  calibration of the statistics between the sample-generating old policies and the policy to update is required through IS weight multiplication \cite{degris2012off}, but using old samples makes large IS weights and this causes large variances in the empirical computation of the loss function. Hence, TRPO and PPO do not consider ER, and ACER  uses clipped IS weights with an episodic replay to avoid large variances, and corrects the bias generated from the clipping \cite{wang2016sample}.

In this paper, we consider  the performance improvement for IS-based PG methods by reusing old samples appropriately based on IS weight analysis and propose a new adaptive multi-batch experience replay (MBER) scheme for PPO, which is currently one of the most popular IS-based PG algorithms. PPO applies clipping but ignores bias, since it  uses the sample batch (or horizon) only from the current policy without replay and hence the required IS weight is not so high. On the contrary to PPO, the proposed scheme uses the batch samples of  past policies as well as the current policy for the update for the next policy, and applies proper techniques  to preserve most advantages of PPO  such as random mini-batch sampling and small bias due to low IS weights. The details of the proposed algorithm will be explained in coming sections.

\vspace{0.5em} {\em Notations:}     $X \sim P$  means that a random variable $X$ follows a probability distribution $P$.  $\Nc(\mu,\sigma^2)$ denotes the Gaussian distribution with mean $\mu$ and variance $\sigma^2$.  ${\mathbb{E}}[\cdot]$ denotes the expectation operator. $\tau_t$ denotes a state-action trajectory from time step $t$: $(s_t,~a_t,~s_{t+1},~a_{t+1},\cdots)$.

\section{Background} \label{sec:Background}

\subsection{Reinforcement Learning Problems} \label{subsec:RLP}

In this paper, we assume that the environment is an MDP. $<\Sc,~\Ac,~\gamma,~P,~r>$ defines a discounted MDP, where $\Sc$ is the state space, $\Ac$ is the action space, $\gamma$ is the discount factor, $P$ is the state transition probability, and $r$ is the reward function.  For every time step $t$, the agent chooses an action $a_t$ based on the current state $s_t$ and then the environment gives the next state $s_{t+1}$ according to $P$  and the reward $r_t=r(s_t,a_t)$ to the agent. Reinforcement learning aims to learn the agent's policy $\pi(a_t|s_t)$ that maximizes  the average discounted return $J=\mathbb{E}_{\tau_0 \sim \pi}[\sum_{t=0}^\infty \gamma^t r_t]$.

\subsection{Deep Q-Learning and PG Methods} \label{subsec:QPGM}

Q-learning is a widely-used reinforcement learning algorithm based on the state-action value function (Q-function).  The state-action value function  represents the expected return of a state-action pair $(s_t,a_t)$ when a policy $\pi$ is used, and is denoted by  $Q_{\pi}(s_t,a_t)=\mathbb{E}_{\tau_t \sim {\pi}}[\sum_{l=t}^\infty \gamma^l r_l]$ \cite{sutton1998reinforcement}. To learn the environment with a discrete action space, DQN approximates the Q-function by using a Q-network $Q_w(s_t,a_t)$ parameterized by $w$, and defines the deterministic policy $\pi(a|s) = \arg\max_{a \in \Ac} Q_w(s,a)$. Then, DQN updates the Q-network parameter $w$ to minimize the temporal difference (TD) error:
\begin{equation}\label{eq:TDerror}
(r_t + \gamma\max_{a'} Q_{w'} (s_{t+1},a') - Q_w(s_t,a_t))^2,
\end{equation}
where $w'$ is the target network. \eqref{eq:TDerror} is from the result of the value iteration algorithm which finds an optimal policy by using the Bellman equation \cite{mnih2013playing}.
Note that the TD error requires maximum of Q-function, but it cannot be computed in a continuous action space. To learn a continuous action environment, PG directly parameterizes the policy by a stochastic policy network $\pi_\theta(a_t|s_t)$ with parameter $\theta$ and sets an objective function $L(\theta)$ to optimize the policy: Value-based PG methods set $L(\theta)$ as the policy follows some distribution of Q-function \cite{lillicrap2015continuous,fujimoto2018addressing,haarnoja2018soft}, and  IS-based PG methods set $L(\theta)$ as the discounted return and directly updates the policy to maximize $L(\theta)$ \cite{schulman2015trust,schulman2017proximal}.


\subsection{IS-based PG and PPO} \label{subsec:PPO}

At each iteration, IS-based PG such as ACER and simple PPO\footnote{We only consider simple PPO without adaptive KL penalty since simple PPO has the best performance on continuous action control tasks.} tries to obtain a better policy  $\pi_{\tilde{\theta}}$ from the current policy $\pi_\theta$ \cite{schulman2015trust}:
\begin{align}\label{eq:ISPG}
L_\theta(\tilde{\theta})&\triangleq\mathbb{E}_{s_t\sim\rho_{\pi_{\theta}},~a_t\sim\pi_{\tilde{\theta}}}
\left[A_{\pi_\theta}(s_t,a_t)\right]\nonumber\\
&=\mathbb{E}_{s_t\sim\rho_{\pi_{\theta}},~a_t\sim\pi_{\theta}}
\left[R_t(\tilde{\theta})A_{\pi_\theta}(s_t,a_t)\right],
\end{align}
where $A_{\pi_{\theta}}(s_t,a_t)=Q_{\pi_{\theta}} (s_t,a_t)-V_{\pi_{\theta}}(s_t)$ is the advantage function, $V_{\pi}(s_t)=\mathbb{E}_{a_t,\tau_{t+1} \sim {\pi}}[\sum_{l=t}^\infty \gamma^l r_l]$ is the state-value function, and $R_t(\tilde{\theta})=\frac{\pi_{\tilde{\theta}}(a_t|s_t)}{\pi_{\theta}(a_t|s_t)}$ is the IS weight.
Here, the objective function $L_\theta(\tilde{\theta})$ is a function of $\tilde{\theta}$ for given $\theta$, and $\tilde{\theta}$ is the optimization variable. To compute  $L_\theta(\tilde{\theta})$ empirically from the samples from the current policy $\pi_\theta$, the IS weight is multiplied. That is, with $R_t(\tilde{\theta})$ multiplied to $A_{\pi_\theta}(s_t,a_t)$, the second expectation in
\eqref{eq:ISPG} is over the trajectory generated by the current policy $\pi_\theta$ not by the updated policy $\pi_{\tilde{\theta}}$.  Here,  large IS weights cause large variances in \eqref{eq:ISPG}, so ACER and PPO bound or clip the IS weight \cite{schulman2017proximal,wang2016sample}. In this paper, we use the clipped important sampling structure of PPO as our baseline. The objective function with clipped IS weights becomes
\begin{equation}\label{eq:PPO}
L_{CLIP}(\tilde{\theta})=\mathbb{E}_{s_t\sim\rho_{\pi_{\theta}},~a_t\sim\pi_{\theta}}
\left[\min\{R_t(\tilde{\theta})\hat{A}_t,~\mathrm{clip}_\epsilon(R_t(\tilde{\theta}))\hat{A}_t\}\right],
\end{equation}
where $\mathrm{clip}_\epsilon(\cdot) = \max(\min(\cdot,1+\epsilon),1-\epsilon)$ with clipping factor $\epsilon$, and $\hat{A}_t$ is the sample advantage function estimated by the generalized advantage estimator (GAE) \cite{schulman2015high}:
\begin{equation}\label{eq:Adv}
\hat{A}_{t} = \sum_{l=0}^{N-n-1} (\gamma\lambda)^{l}\delta_{t+l},
\end{equation}
where $N$ is the number of samples in one iteration (horizon), $\delta_t = r_t + \gamma V_w(s_{t+1}) - V_w(s_t)$ with the state-value network $V_w(s_t)$ which approximates $V_{\pi_{\theta}}(s_t)$. Then, PPO updates the state-value network to minimize the square loss:
\begin{equation}\label{eq:VPPO}
L_V(w) = (V_w(s_t)-\hat{V}_t)^2,
\end{equation}
where $\hat{V}_t$ is the TD($\lambda$) return \cite{schulman2017proximal}.

In \cite{schulman2017proximal}, for continuous action control, a Gaussian policy network is considered, i.e.,
\begin{equation}\label{eq:Gaussian}
 a_t \sim   \pi_{\theta}(\cdot |s_t) =  \mathcal{N}(\mu(s_t;\phi),\sigma^2),
\end{equation}
where $\mu(s_t;\phi)$ is the mean neural network whose input is $s_t$ and parameter is $\phi$; $\sigma$ is a standard deviation parameter; and thus $\theta = (\phi,\sigma)$ is the overall policy parameter.

\section{Related Works} \label{sec:Background}

\subsection{Experience Replay on Q-Learning} \label{subsec:ERQ}

Q-learning is off-policy learning which only requires sampled tuples to compute the TD error \eqref{eq:TDerror} \cite{sutton1998reinforcement}.
DQN uses  the ER technique  \cite{lin1993reinforcement} that stores old sample tuples $(s_t,a_t,r_t,s_{t+1})$ in replay memory $\mathbf{R}$ and updates the Q-network with the average gradient of the TD error computed from a mini-batch uniformly sampled from $\mathbf{R}$. Value-based PG methods adopt this basic ER of DQN.
As an extension of this basic ER, \cite{schaul2015prioritized} considered prioritized ER  to give a sampling distribution on the replay instead of uniform random sampling so that samples with higher TD errors are used more frequently to obtain the optimal Q faster than DQN.  \cite{liu2017effects} analyzed  the effect of the replay memory size on DQN and proposed an adaptive replay memory scheme based on the TD error to find a proper replay size for each discrete task. It is shown that this adaptive replay size for DQN enhances the overall performance.

\subsection{Experience Replay on IS-based PG} \label{subsec:ERPG}

ER can be applied to IS-based PG for continuous action control to increase sample efficiency.  As seen in  \eqref{eq:ISPG},
the multiplication of the IS weight $R_t(\tilde{\theta})=\frac{\pi_{\tilde{\theta}}(a_t|s_t)}{\pi_{\theta}(a_t|s_t)}$ is required to use the samples from old policies for current policy update.
In case that  ER uses samples from many previous policies, the required IS weight is very large and this induces bias even though clipping is applied. The induced bias  makes the learning process  unstable and disturbs the computation of the expected loss function. Thus, ACER uses ER with bias correction, and proposes an episodic ER scheme that samples and stores  on the basis of episodes because it computes an off-policy correction Q-function estimator which requires whole samples in a trajectory as Algorithm 3 in \cite{wang2016sample}.

\section{Multi-Batch Experience Replay} \label{sec:MBER}

\subsection{Batch Structures of ACER and PPO} \label{subsec:ConceptMBER}

Before introducing our new replay scheme, we compare the batch description of ACER and PPO for updating the policy, as shown in Fig. \ref{fig:structComp}. ACER uses an episodic ER to increase sample efficiency. In the continuous action case, ACER stores trajectories from $V=100$ previous policies and each policy generates a trajectory of $M=50$ time steps in replay memory $\Rbf$. For each update period, ACER chooses $W \sim Poisson(4)$ random episodes from $\Rbf$ to update the policy. Then, different statistics among the samples in the replay causes bias, and the episodic sample mini-batch is highly correlated. On the other hand,  PPO does not use ER but collects a single batch of size $N=2048$ time steps from the current policy. Then, PPO  draws a mini-batch of size  $M=64$ randomly and uniformly from the single batch; updates the policy to the direction of the gradient of the empirical loss computed from the drawn mini-batch:
\begin{equation}\label{eq:PPOemp}
\hat{L}_{CLIP}(\tilde{\theta})=\frac{1}{M}\sum_{m=0}^{M-1}\min\{R_m(\tilde{\theta})\hat{A}_m,~\mathrm{clip}_\epsilon(R_m(\tilde{\theta}))\hat{A}_m\};
\end{equation}
and updates the value network to the direction of the negative gradient of
\begin{equation}\label{eq:VPPOemp}
\hat{L}_V(w)=\frac{1}{M}\sum_{m=0}^{M-1}(V_w(s_m)-\hat{V}_m)^2,
\end{equation}
where $\hat{A}_m$, $R_m(\tilde{\theta})$, $V_w(s_m)$, and $\hat{V}_m$ are values corresponding to the $m$-th sample in the mini-batch \cite{schulman2017proximal}. This procedure is repeated
for $10$ epochs for a single batch of size $N$.\footnote{1 epoch means that we use every samples in the batch to update it once. In other words, PPO updates the policy by drawing $10 \cdot N/M$ mini-batches.}
 Note that PPO uses current samples only, so it can ignore  bias because the corresponding IS weights do not exceed the clipping factor mostly. Furthermore, the samples in a mini-batch drawn uniformly from the total batch of size $N$ in PPO have little sample correlation because they are scattered over the total batch. However, PPO discards all samples from all the past policies except the current policy for the next policy update and this reduces sample efficiency.

\subsection{The Proposed Multi-Batch Experience Replay Scheme} \label{subsec:ConceptMBER}

We now present our  MBER scheme suitable to PPO-style IS-based PG, which increases sample efficiency, maintains random mini-batch sampling to diminish the sample correlation, and reduces the IS weight to avoid bias.
We apply our MBER scheme to PPO to construct an enhanced algorithm named PPO-MBER, which includes PPO as a special case.

\begin{figure*}[!h]
	\centering
	\includegraphics[width=0.95\textwidth]{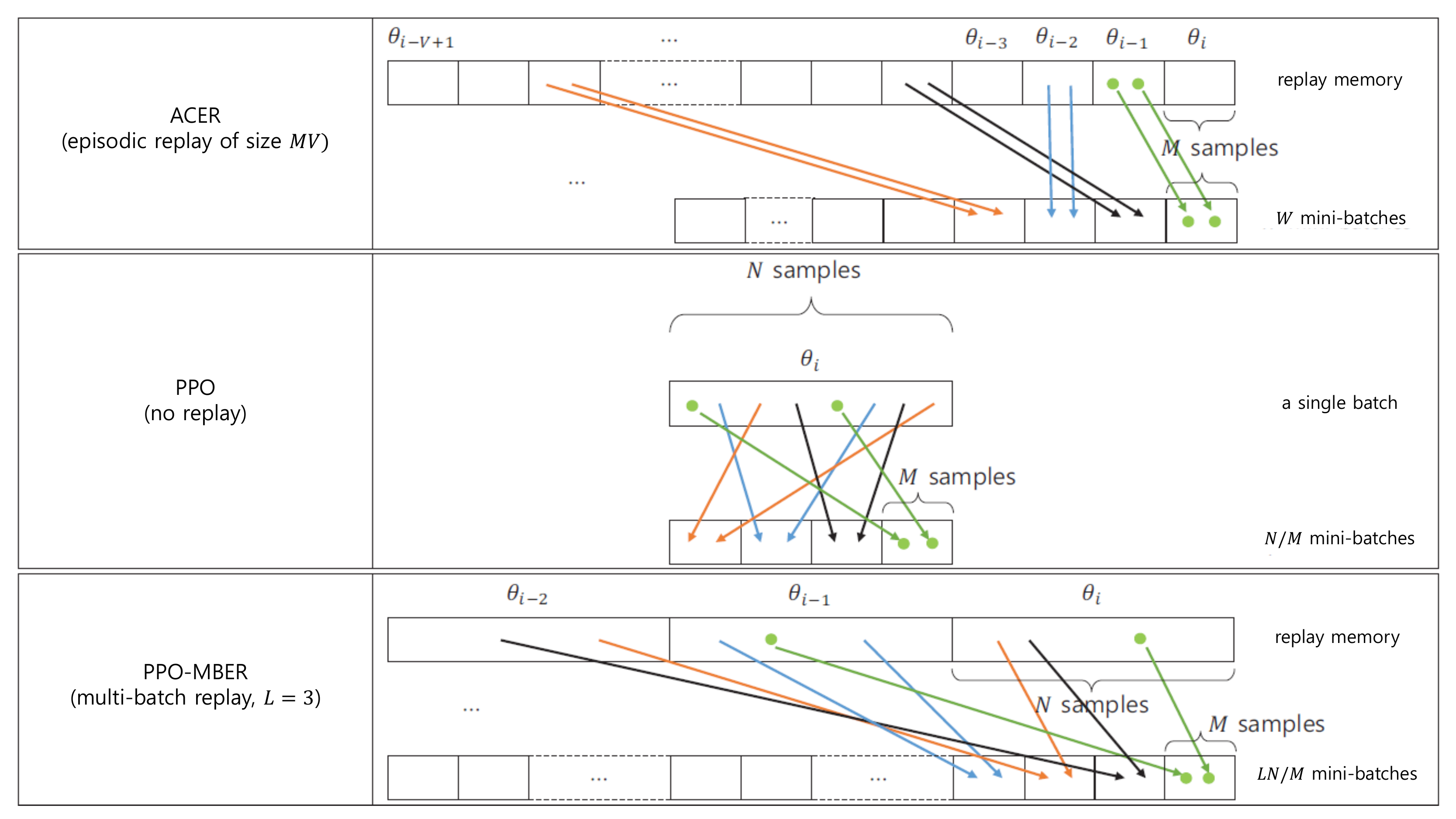}
	\caption{Batch construction of ACER, PPO and PPO with the proposed MBER (PPO-MBER): $N=8$ and $M=2$}
	\label{fig:structComp}
\end{figure*}

In order to obtain the next policy, the proposed scheme uses the batch samples of $L-1$ past policies and the current policy, whereas original PPO uses the batch samples from only the current policy, as illustrated in Fig. \ref{fig:structComp}.
The stored information for MBER in the replay memory $\Rbf$ is as follows. To compute the required IS weight $R_t(\tilde{\theta})=\frac{\pi_{\tilde{\theta}}(a_t|s_t)}{\pi_{\theta}(a_t|s_t)}$ for each sample in a random mini-batch,
 MBER stores the statistical information of every sample in $\Rbf$. Under the assumption of a Gaussian policy network \eqref{eq:Gaussian}, the required statistical information for each sample is
  the mean $\mu_t := \mu(s_t;\phi)$ and the standard deviation $\sigma$.  Furthermore, MBER stores the pre-calculated  estimated advantage $\hat{A}_t$ and target value $\hat{V}_t$ of every sample in $\Rbf$.
     Thus, MBER stores the overall sample information  $(s_t,~a_t,~\hat{A}_t,~\hat{V}_t,\mu_t,\sigma)$ regarding the batch samples from the most recent $L$ policies,  as described in Fig. \ref{fig:structComp}.
    The storage of the statistical information $(\mu_t,\sigma)$ and the values $(\hat{A}_t,\hat{V}_t)$ in addition to $(s_t,a_t)$ for every sample in the replay memory makes it possible
  to draw  a random mini-batch from $\Rbf$ not a trajectory like in ACER.
Since the policy at the $i$-th iteration generates a batch of $N$ samples, we can rewrite the stored information by using
two indices $i=0,1,\cdots$ and $n=0,1,\cdots,N-1$ (such that time step $t=iN+n$) as $\{B_{i-L+1},\cdots,B_i\}$ from the most recent $L$  policies $i,i-1,\cdots,i-L+1$, where
\begin{equation}
B_i = (s_{i,n},~a_{i,n},~\hat{A}_{i,n},~\hat{V}_{i,n},\mu_{i,n},\sigma_i),~~~n=0,\cdots,N-1.
\end{equation}

\begin{figure}[!h]
	\centering
	\includegraphics[width=0.4\textwidth]{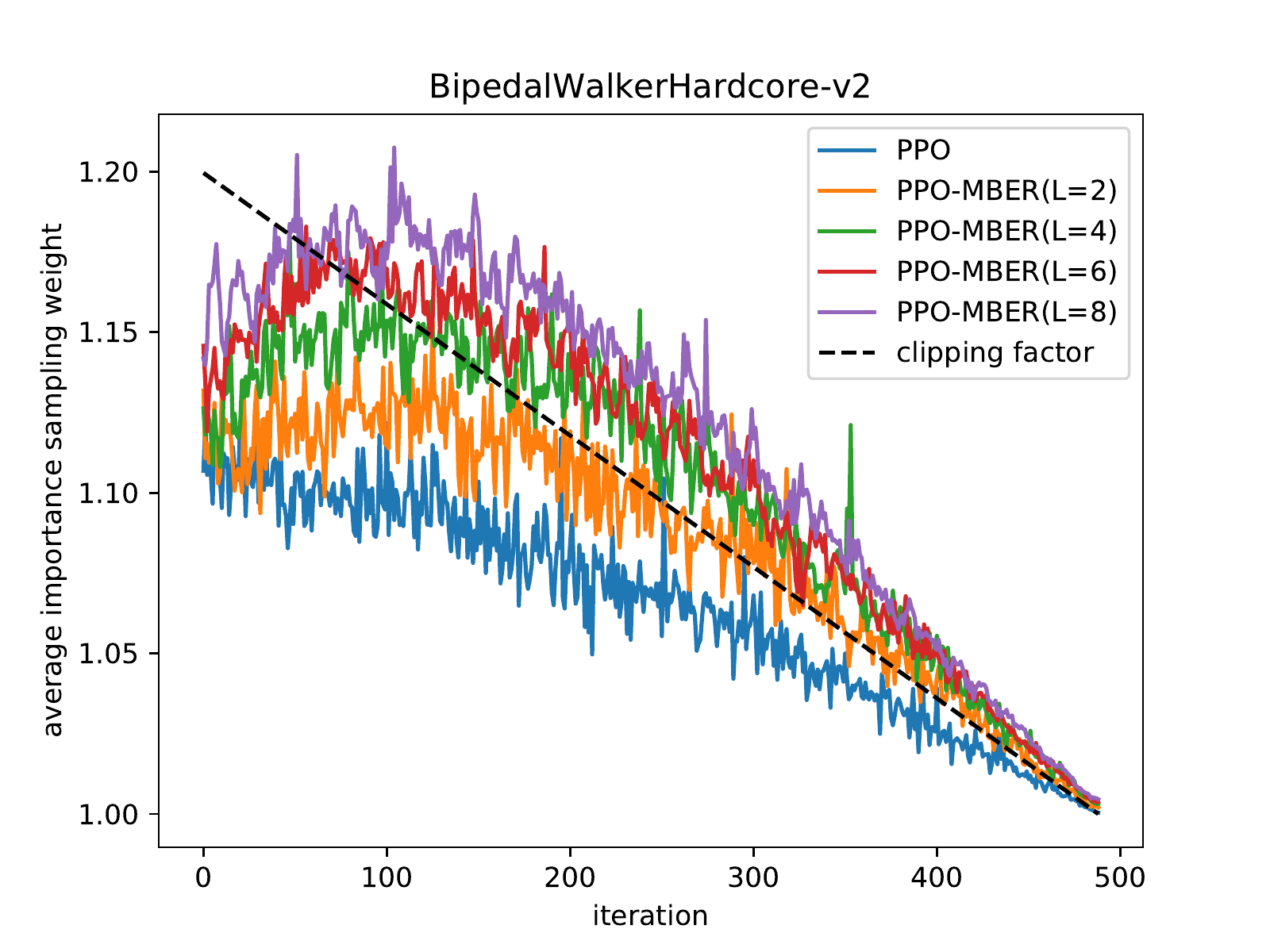}
	\includegraphics[width=0.4\textwidth]{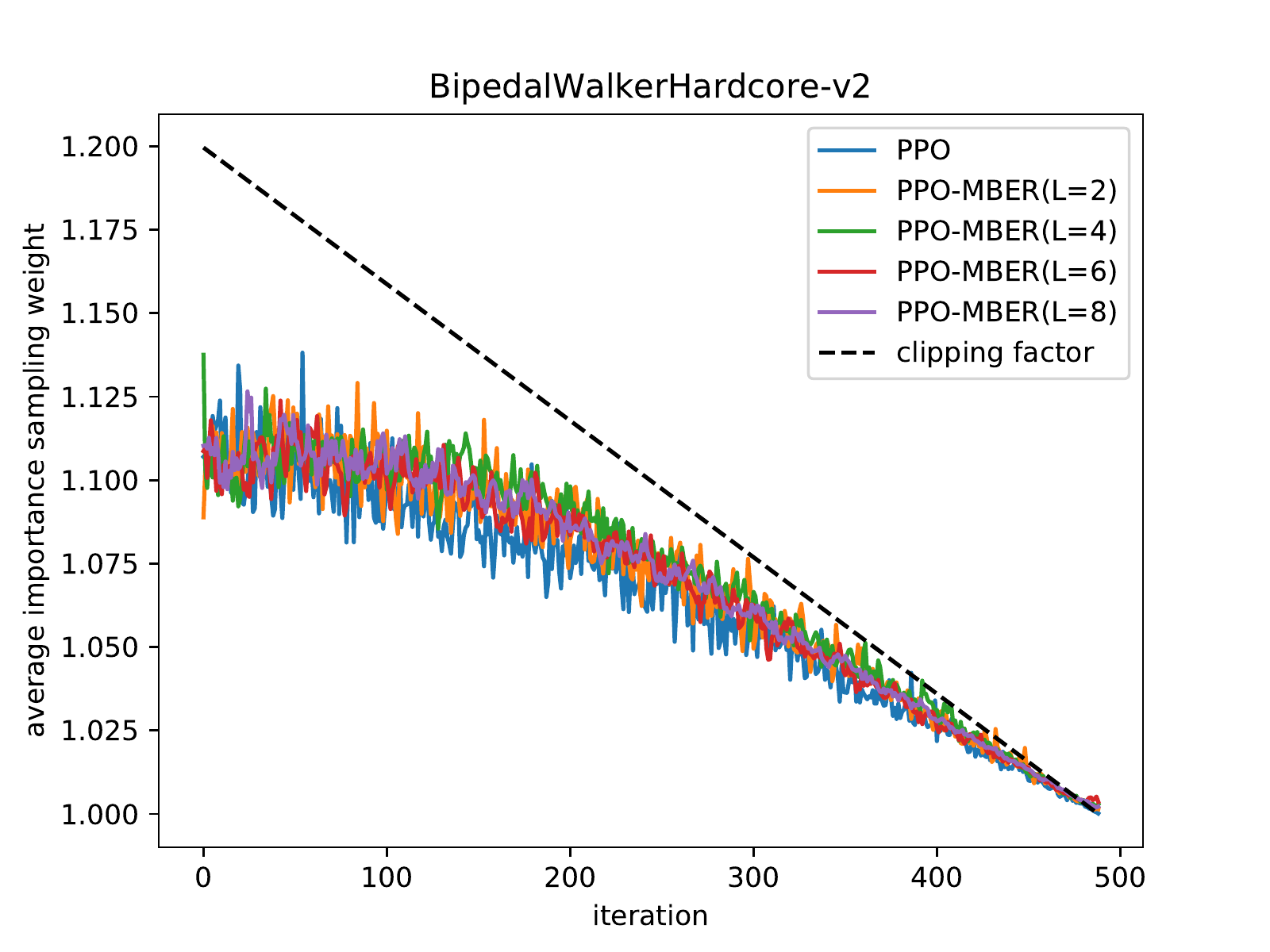}
	\caption{Average IS weight of BipedalWalkerHardcore for PPO-MBER: (Upper) $M=64,~\epsilon=0.2$  and (Lower) $M=64L,~\epsilon=0.2$}
	\label{fig:ImpComp}
\end{figure}

In addition to using the batch samples from most recent $L$ policies, MBER enlarges the mini-batch size by $L$ times compared to that of original PPO, to reduce the average IS weight. If we set the mini-batch size of MBER to be the same as that of PPO with the same epoch, then the number of updates of PPO-MBER is $L$ times larger than that of PPO. Then, the updated policy statistic is too much different from the current policy statistic and thus the average IS weight becomes large as $L$ increases. This causes bias and is detrimental to the performance. To avoid this, we enlarge the mini-batch size of MBER by $L$ times and this reduces the average IS weight by making the number of updates the same as that of PPO with the same epoch. In this way, MBER can ignore bias without much concern, because its IS weight is similar to that of PPO.  Fig. \ref{fig:ImpComp} shows  the average IS weight\footnote{Actually, we averaged $\mathrm{abs}(1-R_m(\tilde{\theta}))+1$ instead of $R_m(\tilde{\theta})$ to see the degree of deviation from $1$.} of all sampled mini-batches at each iteration when $M=64$ and $M=64L$.  It is seen  that PPO-MBER maintains the same level of the important sampling weight as original PPO.

\section{Adaptive Batch Drop} \label{sec:Adaptive}

\begin{figure}[!h]
	\centering
	\includegraphics[width=0.32\textwidth]{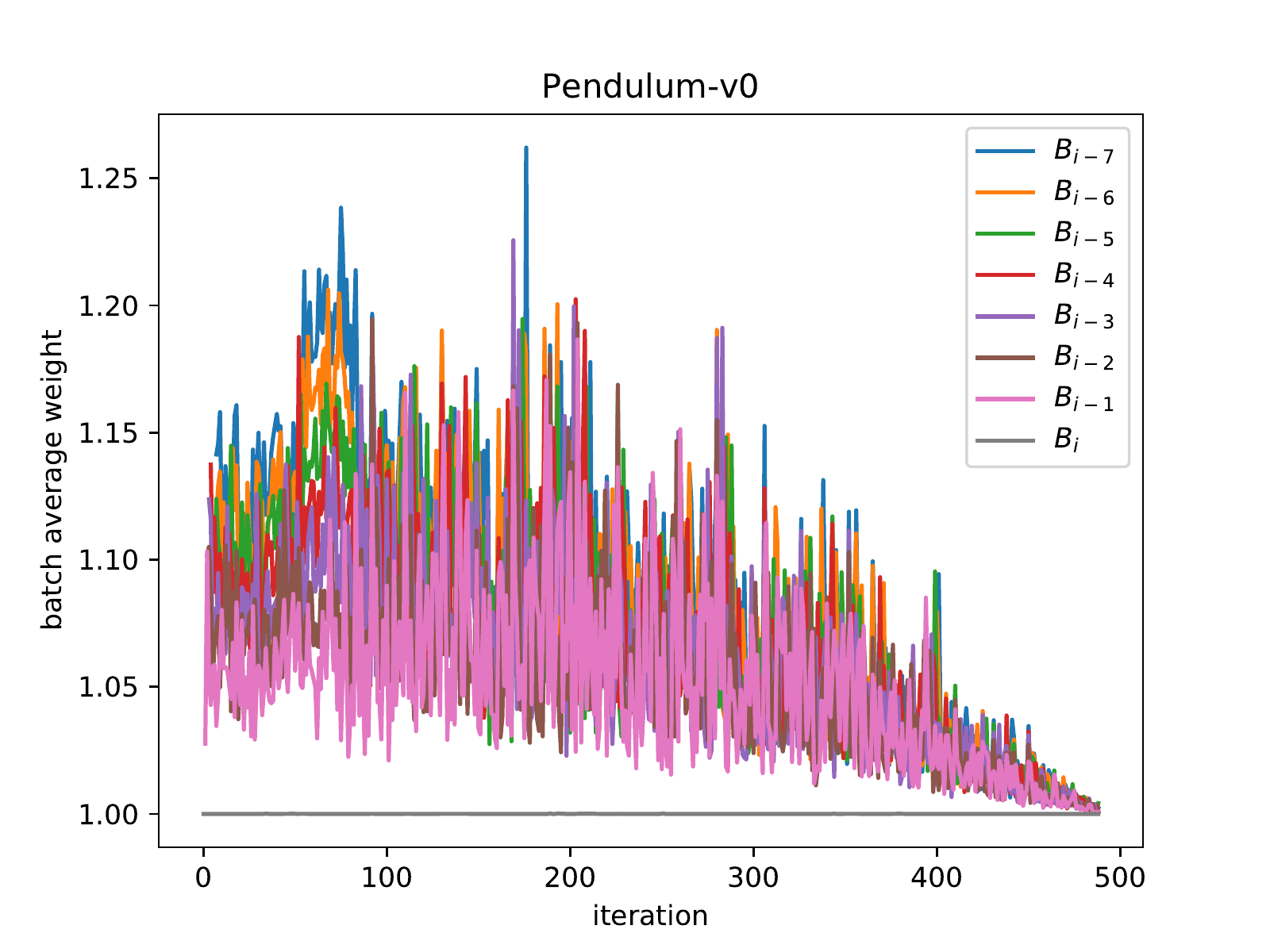}
	\includegraphics[width=0.32\textwidth]{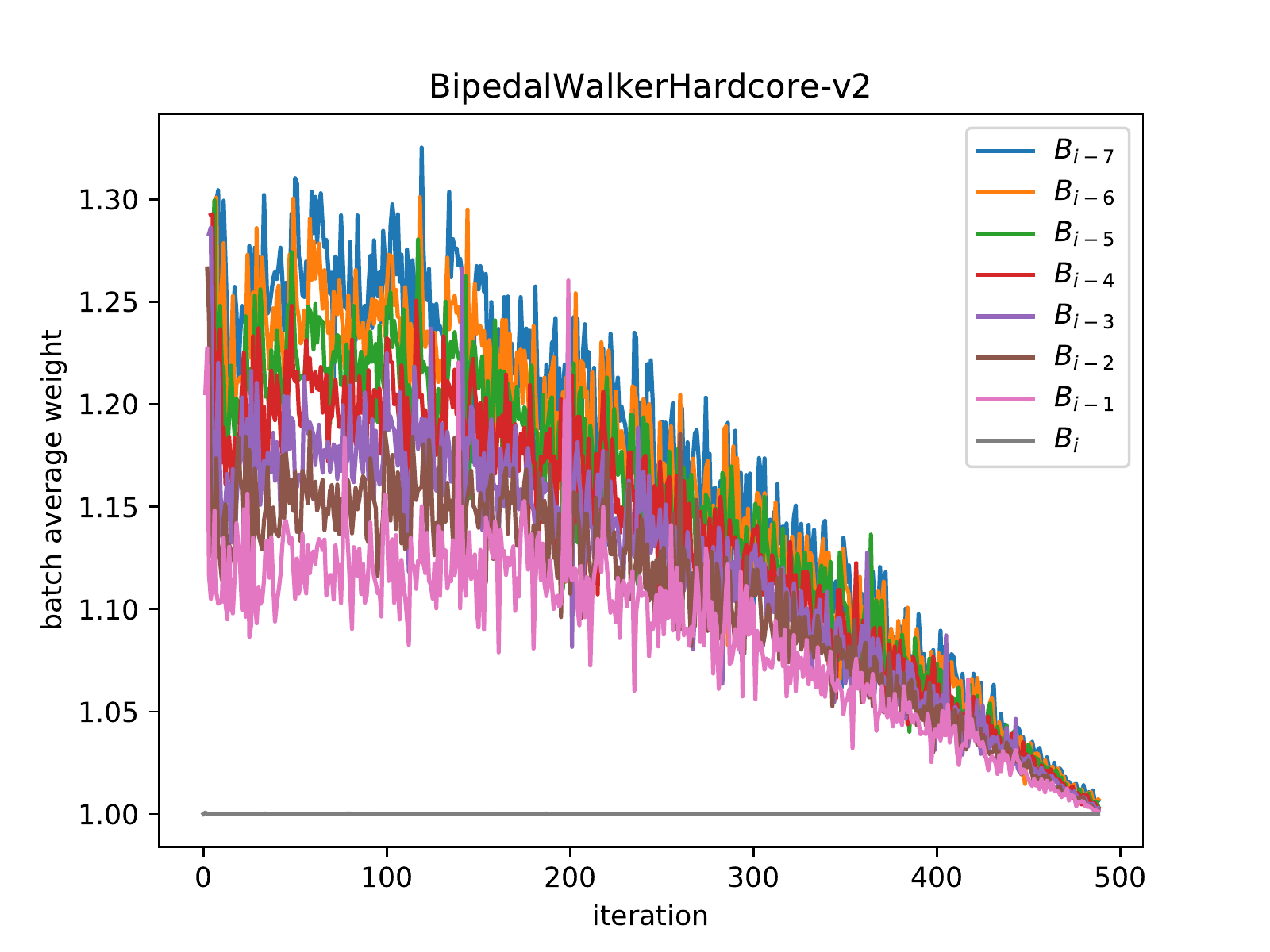}
	\includegraphics[width=0.32\textwidth]{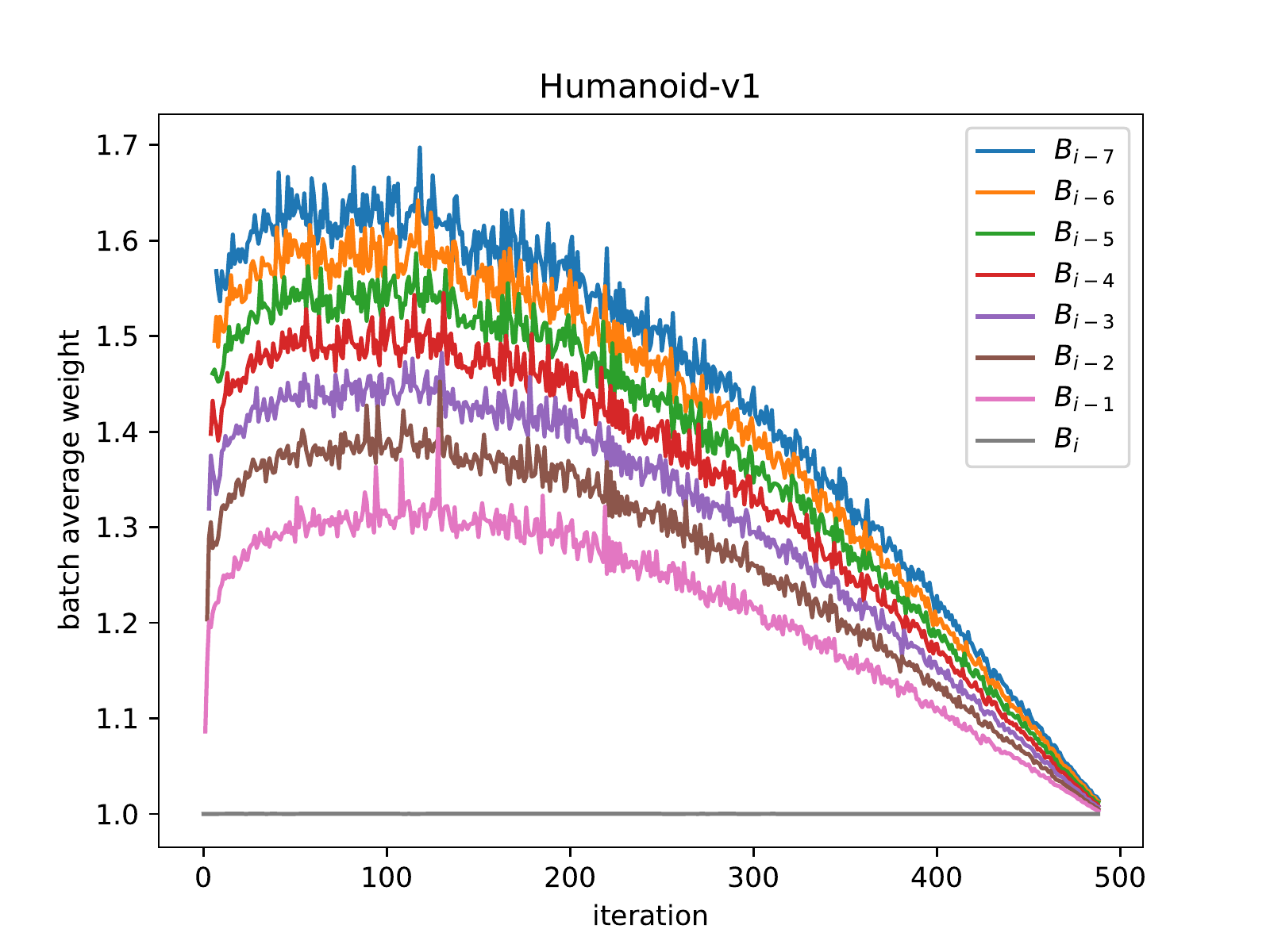}
	\caption{Batch average weight $R'_{i,l}$ of PPO-MBER ($L=8,~\epsilon=0.4$): (Upper) Pendulum, (Center) BipedalWalkerHardcore, and (Lower) Humanoid}
	\label{fig:BatchAvgWeight}
\end{figure}

PPO-MBER  can significantly enhance the overall performance compared to PPO by using the MBER  scheme, as seen in Section \ref{sec:Experiments}. However, we observe that the PPO-MBER performance for each task  depends on the replay length $L$, and hence the choice of $L$ is crucial to PPO-MBER. For the two extreme examples, Pendulum and Humanoid, in Table \ref{table:LastResult}, note that the performance of Pendulum is proportional to the replay length but the performance of Humanoid is inversely proportional to the replay length. To analyze the cause of this phenomenon, we define the batch average IS weight between the old policy $\theta_{i-l}$ and the current policy $\theta_i$ as
\begin{equation}
R'_{i,l} = \frac{1}{N}\sum_{n=1}^{N-1}1+\mathrm{abs}\left(1-\frac{\pi_{\theta_i}(a_{i-l,n}|s_{i-l,n})}{\pi_{\theta_{i-l}}(a_{i-l,n}|s_{i-l,n})}\right),
\end{equation}
where $a_{i-l,n},~s_{i-l,n}\in B_{i-l}$.
Note that this  is different from the average of $1+\mathrm{abs}(1-R_m(\tilde{\theta}))$ which depends on the updating policy $\pi_{\tilde{\theta}}$ not the current policy $\pi_{\theta_i}$.
Fig. \ref{fig:BatchAvgWeight} shows $R'_{i,l}$ of PPO-MBER for Pendulum, Humanoid, and BipedalWalkerHardcore tasks.
It is seen that  $R'_{i,l}$ increases as $l$ increases, because the batch statistic is updated as iteration goes on. It is also seen that Humanoid has "large"  $R'_{i,l}$  and Pendulum has "small" $R'_{i,l}$. From the two examples, it can be inferred that
the batch samples  $B_{i-l}$ with large $R'_{i,l}$  are too old for updating $\tilde{\theta}$ at the current policy parameter $\theta_i$ and can harm the performance, as in the case of Humanoid. On the other hand, if $R'_{i,l}$ is small, more old samples can be used for update and this is beneficial to the performance.  Therefore, it is observed in Table \ref{table:LastResult} that original PPO, i.e., $L=1$ is best for Humanoid and PPO-MBER with $L=8$ is best for Pendulum (In Table \ref{table:LastResult}, we only consider $L$ up to 8). Exploiting this fact,
 we propose an adaptive MBER (AMBER) scheme which adaptively chooses the batches to use for update from the replay memory. In the proposed AMBER, we store the batch samples from policies $\theta_i,\theta_{i-1},\cdots,\theta_{i-L+1}$, but use only the batches $B_{i-l}$'s whose $R'_{i,l}$ is smaller than the batch drop factor $\epsilon_b$.  Since $R'_{i,l}$ increases as time goes, AMBER uses the most recent $L'$ sample batches whose $R'_{i,l}$ is less than $\epsilon_b$. It is seen in Table \ref{table:LastResult} that PPO-AMBER well selects the proper replay length.

\section{The Algorithm} \label{sec:Algorithm}

Now, we present our proposed algorithm PPO-(A)MBER that maximizes the objective function $\hat{L}_{CLIP}(\tilde{\theta})$ in \eqref{eq:PPOemp} for continuous action control. We assume the Gaussian policy network $\pi_{\tilde{\theta}}$ in \eqref{eq:Gaussian} and the value network $V_w$. (They do not share parameters.) We define the overall parameter $\tilde{\theta}_{ALL}$ combining the policy parameter $\tilde{\theta}$ and the value parameter $w$. The objective function is given by \cite{schulman2017proximal}
\begin{equation}  \label{eq:GeneralTarget}
\hat{L}(\tilde{\theta}_{ALL})=\hat{L}_{CLIP}(\tilde{\theta}) - c_v \hat{L}_V (w),
\end{equation}
where $\hat{L}_{CLIP}(\tilde{\theta})$ is in
\eqref{eq:PPOemp},   $\hat{L}_V (w)$ is in
\eqref{eq:VPPOemp}, and $c_v$ is a constant (we use $c_v=1$ in the paper). The algorithm is summarized as Algorithm \ref{algorithm:PPO-AMBER} in Appendix.

\section{Experiments} \label{sec:Experiments}

\subsection{Environment Description and Parameter Setup} \label{subsec:PerfComp}

To evaluate our ER scheme, we conducted numerical experiments on OpenAI GYM environments \cite{brockman2016openai}. We selected continuous action control environments of GYM: Mujoco physics engines \cite{todorov2012mujoco}, classical control, and Box2D \cite{catto2011box2d}. The dimensions of state and action for each task are described in Table \ref{table:dim}.
\begin{table}[!h]\footnotesize
	\caption{Description of Continuous Action Control Tasks}\label{table:dim}
	\centering
	\begin{tabular}{lC{1.5cm}C{1.5cm}}
		\toprule
		Mujoco Tasks & State dim. & Action dim.\\
		\cmidrule(r){1-3}
		HalfCheetah-v1 & 17 & 6 \\
		Hopper-v1 & 11 & 3 \\
		HumanoidStandup-v1 & 376 & 17 \\
		Humanoid-v1 & 376 & 17 \\
		InvertedDoublePendulum-v1 & 11 & 1 \\
		InvertedPendulum-v1 & 4 & 1 \\
		Swimmer-v1 & 8 & 2 \\
		Reacher-v1 & 11 & 2 \\
		Walker2d-v1 & 17 & 6 \\
		\midrule
		Classic Control & State dim. & Action dim.\\
		\midrule
		Pendulum-v0 & 3 & 1 \\
		\midrule
		Box2D & State dim. & Action dim.\\
		\midrule
		BipedalWalker-v2 & 24 & 4 \\
		BipedalWalkerHardcore-v2 & 24 & 4 \\
		\bottomrule
	\end{tabular}
\end{table}
We used PPO baselines of OpenAI \cite{baselines} and compared the performance of PPO-(A)MBER with various replay lengths $L=1$ (PPO)$, 2, 4, 6, 8$ on continuous action control tasks in Table \ref{table:dim}. The hyperparameters of PPO/PPO-MBER are described in Table \ref{table:PPOparam}: Adam step size $\beta$ and clipping factor $\epsilon$ decay linearly as time-step goes on from the initial values to $0$. The Gaussian mean network and the value network are feed-forward neural networks that have $2$ hidden layers of size $64$ like in \cite{schulman2017proximal}.
For all the performance plots/tables in this paper, we performed $10$ simulations per each task with random seeds.
For each performance plot, the $X$-axis is time step,  the $Y$-axis is the average return of the lastest $100$ episodes at each time step, and the line in the plot is the mean performance of $10$ random seeds.
For each performance table, results are described  as the mean $\pm$ one standard deviation of $10$ seeds  and the best scores are in boldface.
To measure the overall performance of an algorithm on various continuous control tasks, we first compute the normalized score (NS) over all simulation setups
in this paper for each task, then compute the  average NS (ANS) which is the averaged over all tasks like in \cite{schulman2017proximal}. It can be thought that the ANS of final $100$ episodes indicates the performance after convergence and the ANS of all episodes indicates the speed of convergence.
We refer to the former  as the final ANS and the latter as the speed ANS. The ANS results of all simulation settings in this paper are summarized in Appendix.

\subsection{Performance and Ablation Study of PPO-MBER} \label{subsec:AblationClip}

In \cite{schulman2017proximal}, the optimal clipping factor is $0.2$ for PPO. However, it depends on the task set. Since our task set is a bit different from that in \cite{schulman2017proximal}, we first evaluated the performance of PPO and PPO-MBER by sweeping the clipping factor from $\epsilon=0.2$ to $\epsilon = 0.7$, and the corresponding final/speed ANS results are summarized in Tables \ref{table:FinalANS} and \ref{table:FastANS}, respectively.
From the results, we observe that loosening the clipping factor a bit is beneficial for both PPO and PPO-MBER in the considered set of tasks, especially PPO-MBER with larger replay lengths.
This is because loosening the clipping factor reduces the bias and increases the variance of the loss expectation, but a larger mini-batch of PPO-MBER  reduces the variance by offsetting\footnote{One may think that PPO with a large mini-batch size also has the same effect, but enlarging the mini-batch size without increasing the replay memory size increases the sample correlation and reduces the number of updates too much, so it is not helpful for PPO.}. However, too large a clipping factor harms the performance. The best clipping factor is $\epsilon = 0.3$ for PPO and $\epsilon=0.4$ for PPO-MBER.
The detailed score for each task for PPO and PPO-MBER with the best clipping factors is given in Table. \ref{table:LastResult} and \ref{table:AvgResult}.
The PPO results match with those in \cite{schulman2017proximal} for most environments, but note that the Swimmer performance of PPO in our result is a little worse than that of \cite{schulman2017proximal}. This  is because PPO sometimes fails to perfectly learn the environment as the number of random seeds increases from $3$ of \cite{schulman2017proximal} to $10$ of ours.
However, PPO-MBER  learns the Swimmer environment more stably since it averages more samples based on enlarged mini-batches, so there is a large performance gap in this case.
It is observed that in most environments,  PPO-MBER with proper $L$ significantly enhances both the final and speed ANS results as compared to PPO.

\begin{figure}[!h]
	\centering
	\includegraphics[width=0.32\textwidth]{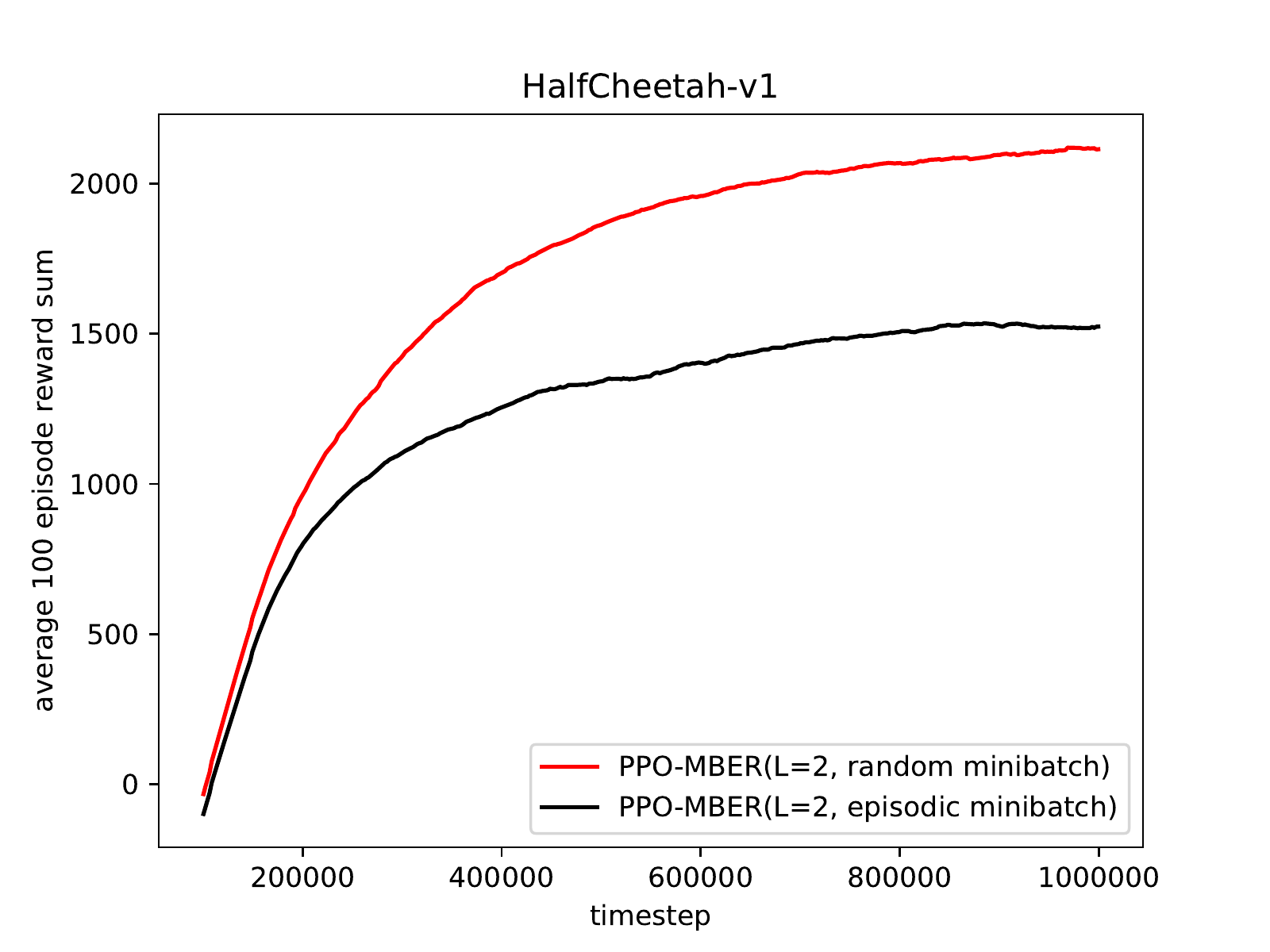}
	\includegraphics[width=0.32\textwidth]{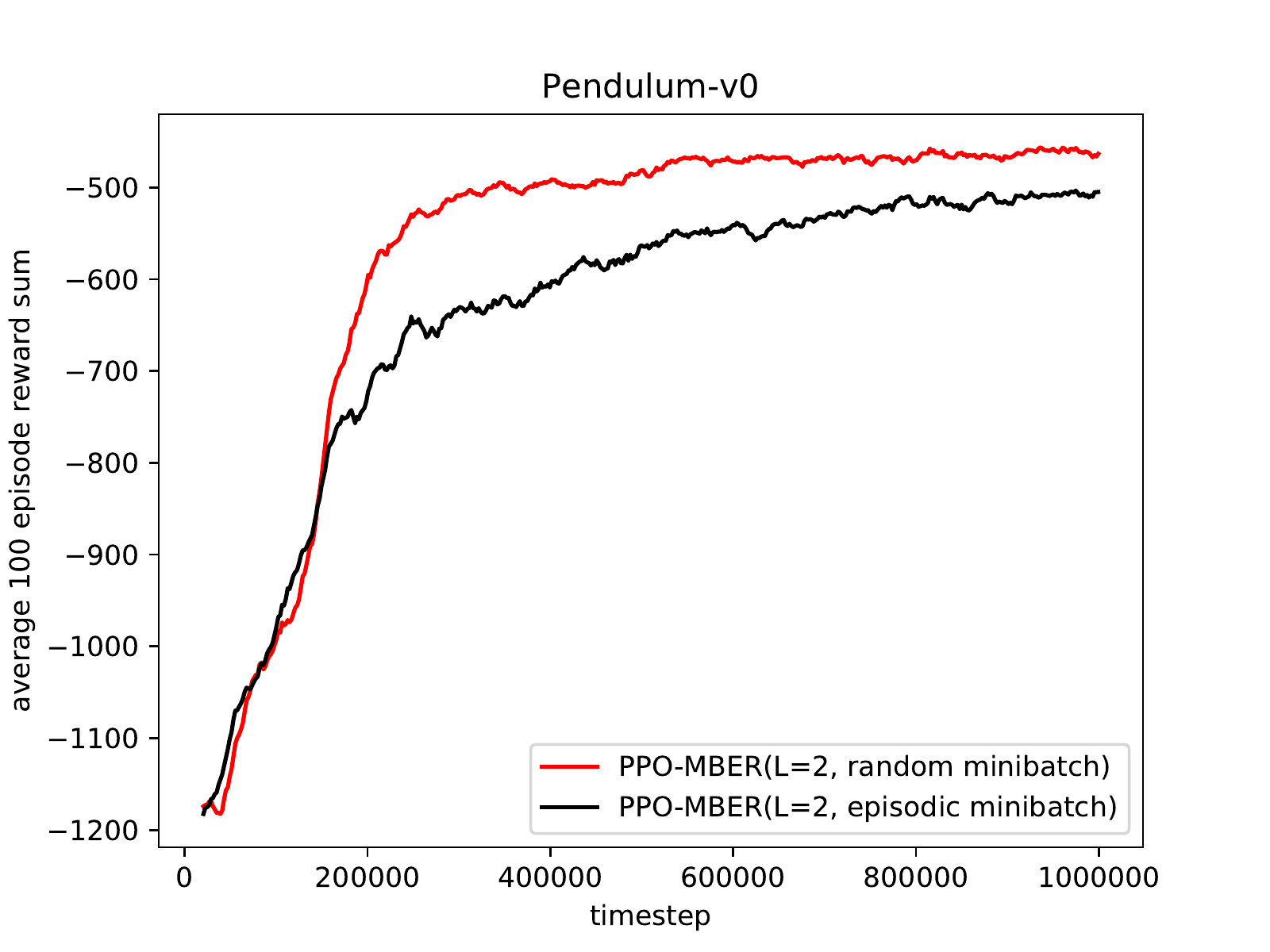}
	\includegraphics[width=0.32\textwidth]{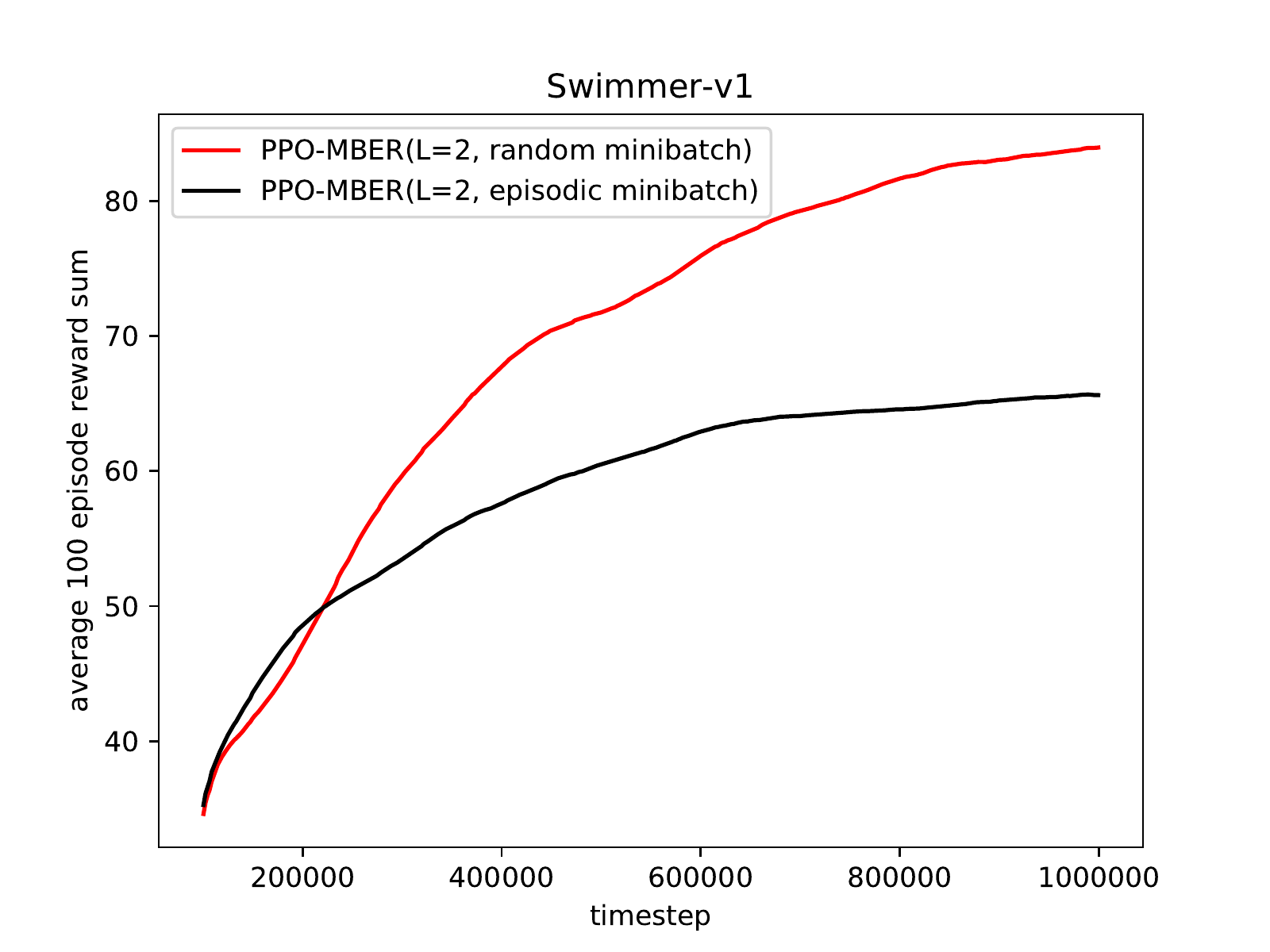}
	\caption{Performance comparison of HalfCheetah, Pendulum, and Swimmer for PPO-MBER $(L=2,~\epsilon=0.4)$ with uniformly random mini-batch and episodic mini-batch}
	\label{fig:EpisodicCompare}
\end{figure}

We then investigated the performance of random mini-batches versus episodic mini-batches for PPO-MBER with $L=2,~\epsilon=0.4$.
In the episodic case, we draw each mini-batch by picking a consequent trajectory of size $M$ like ACER.
Fig. \ref{fig:EpisodicCompare} shows the results under  several environments.
It is seen that there is a notable performance gap between the two cases. This means that random mini-batch drawing from the replay memory storing pre-computed advantages and values in MBER has the advantage of reducing the sample correlation in a mini-batch and this is beneficial to  the performance.

\subsection{Performance of PPO-AMBER} \label{subsec:AdaptivePerf}

\begin{figure}[!h]
	\centering
	\includegraphics[width=0.32\textwidth]{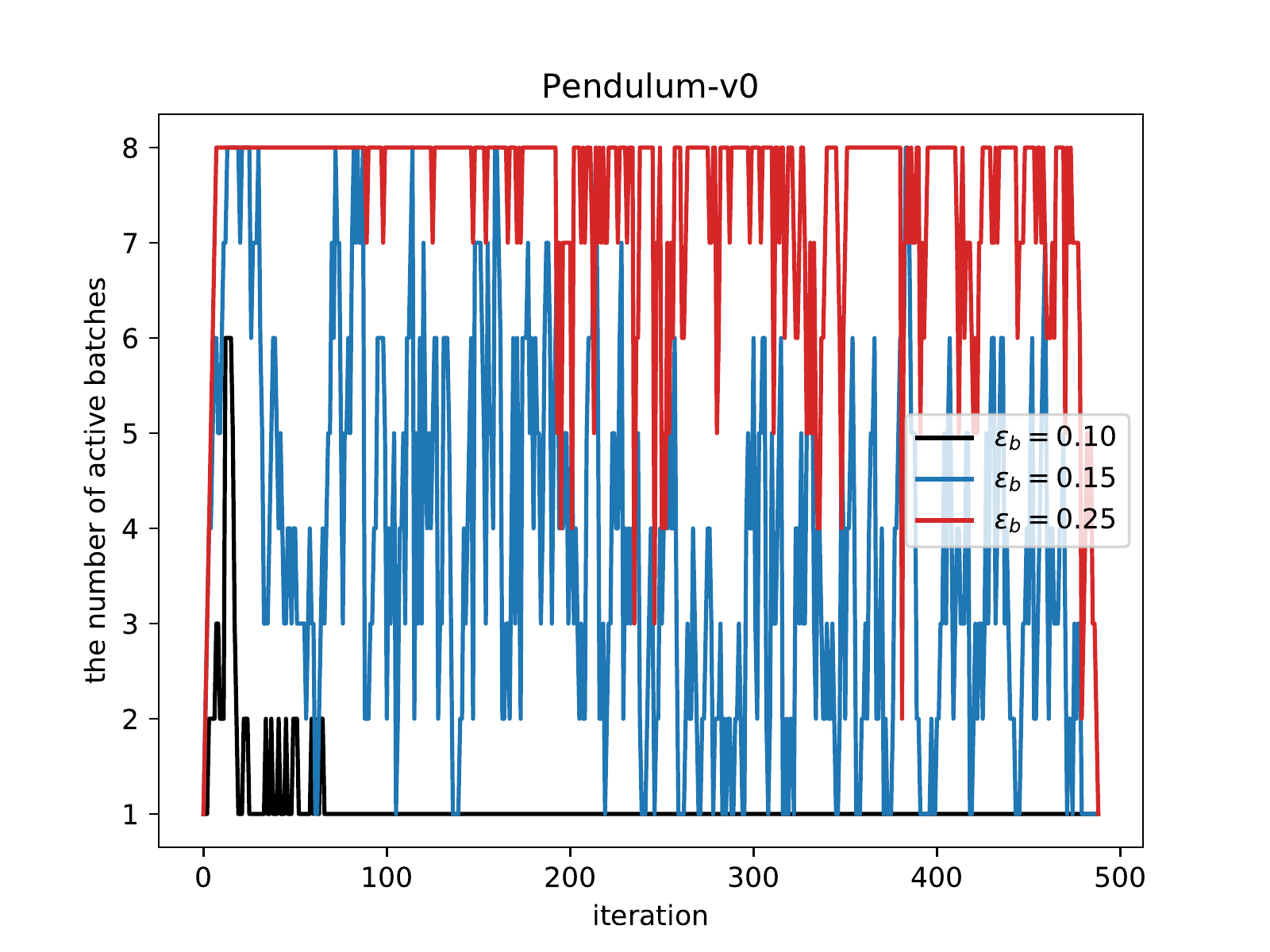}
	\includegraphics[width=0.32\textwidth]{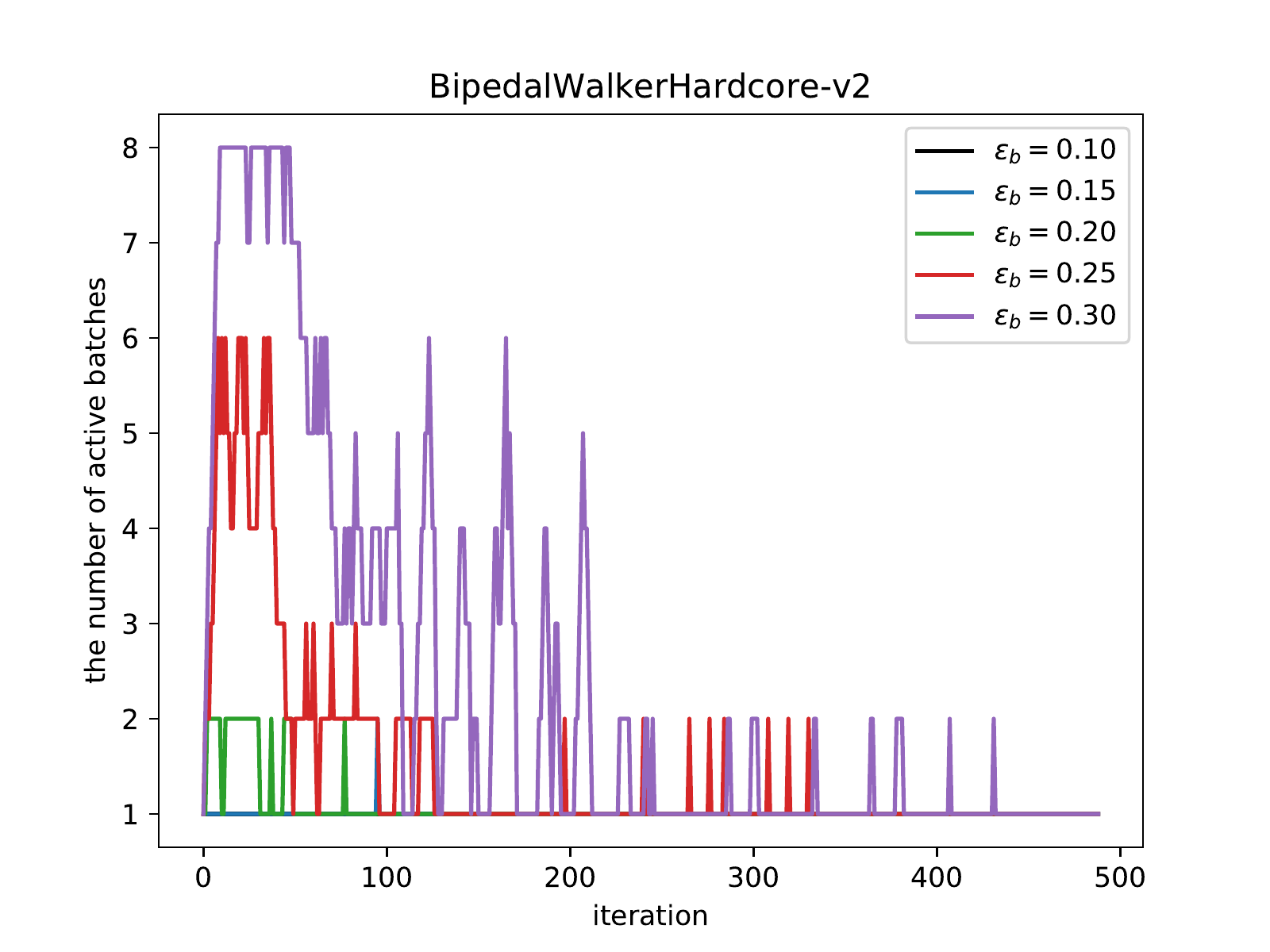}
	\includegraphics[width=0.32\textwidth]{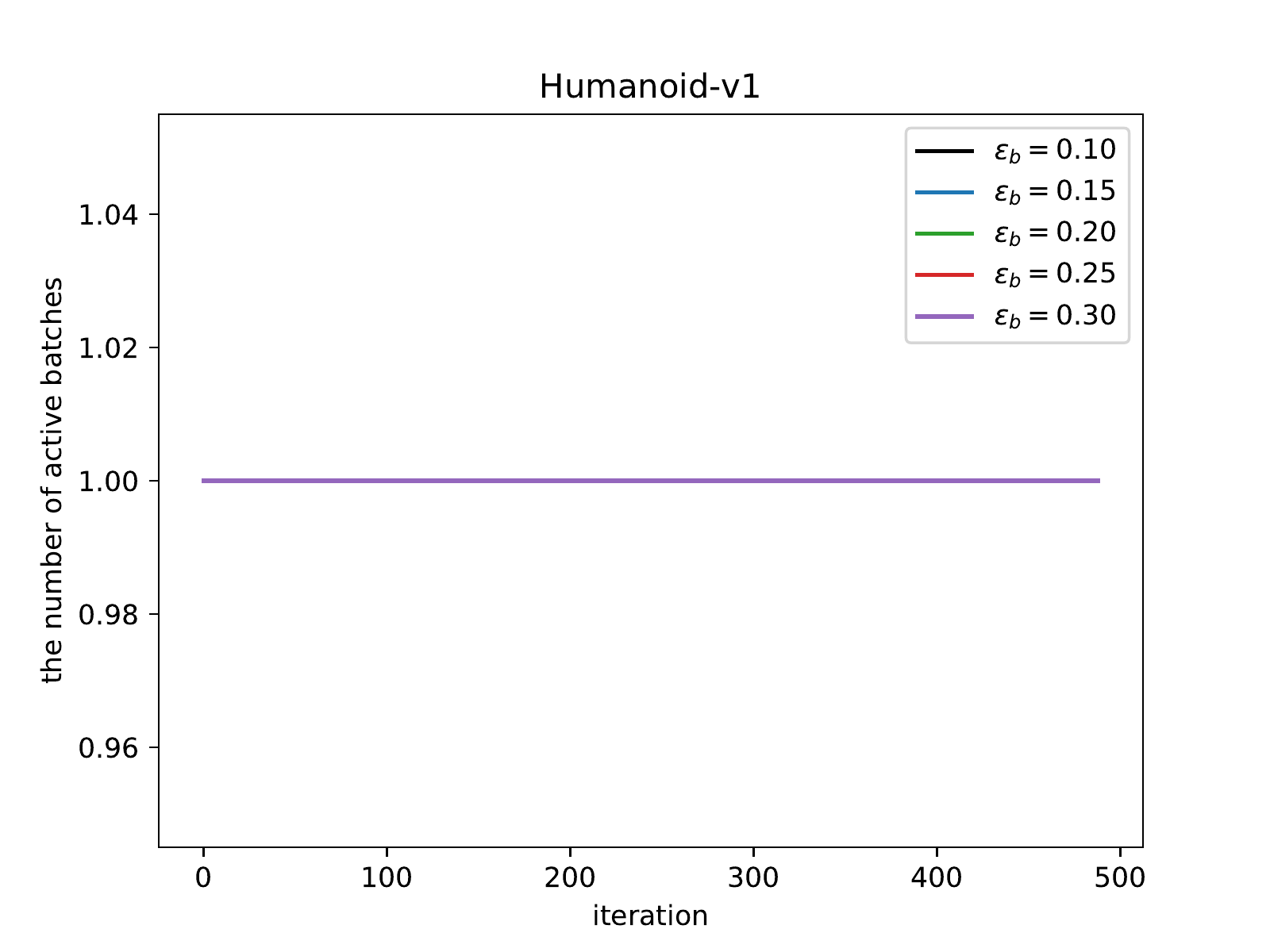}
	\caption{The number of active batches of Pendulum, BipedalWalkerHardcore, and Humanoid for PPO-AMBER with various $\epsilon_b$}
	\label{fig:AdaptiveBatch}
\end{figure}

\begin{figure*}[!h]
	\centering
	\includegraphics[width=0.32\textwidth]{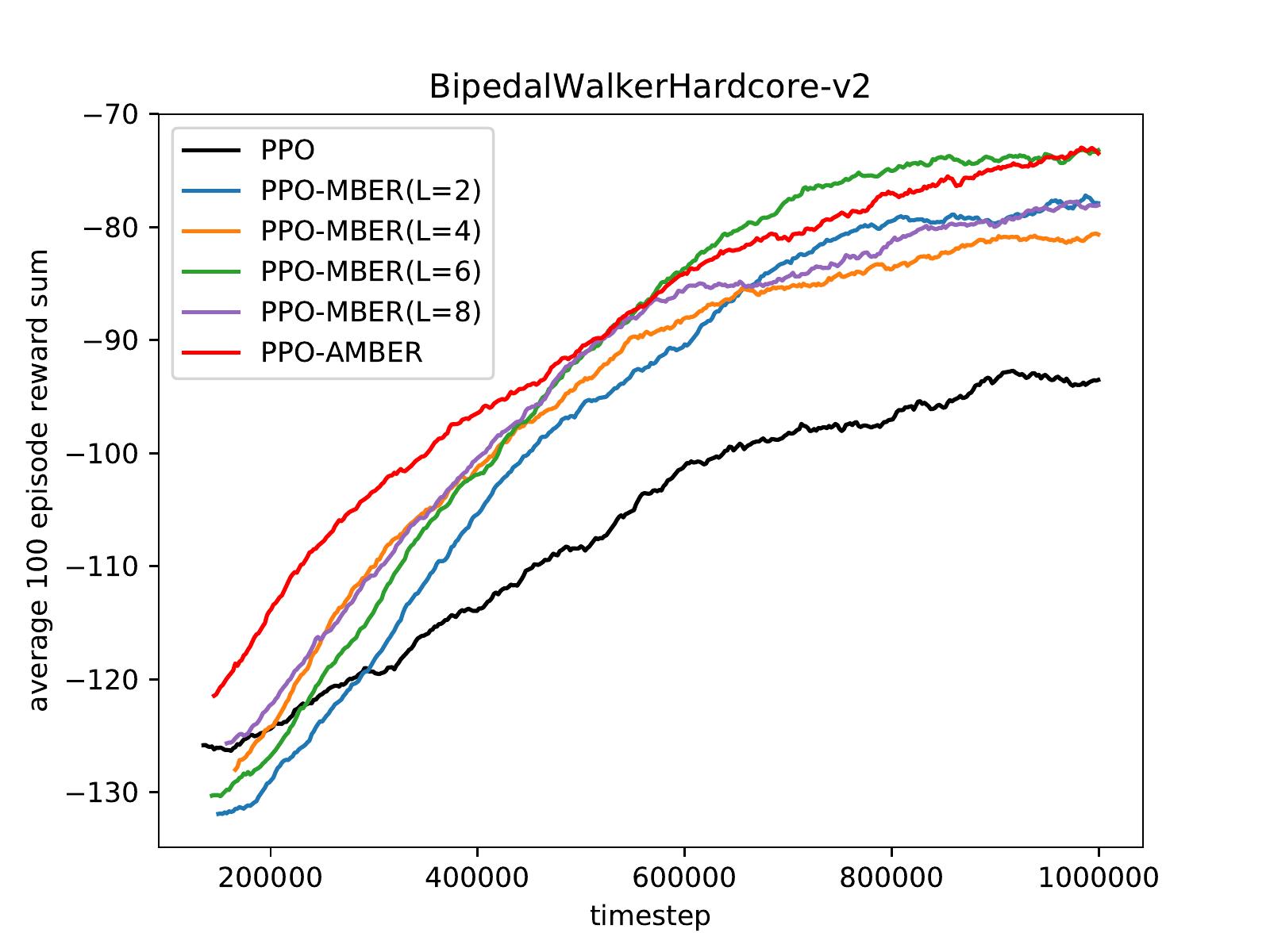}
	\includegraphics[width=0.32\textwidth]{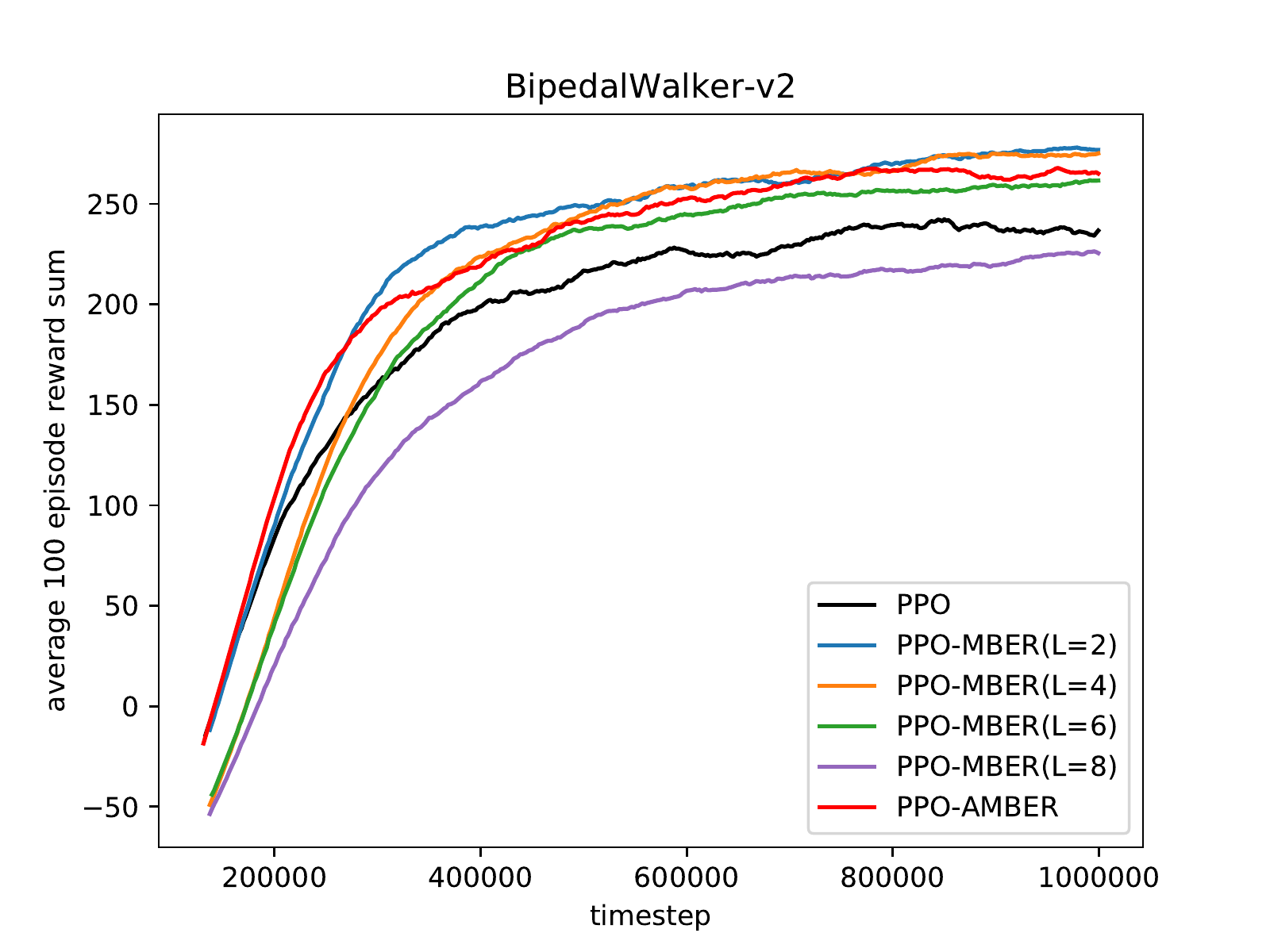}
	\includegraphics[width=0.32\textwidth]{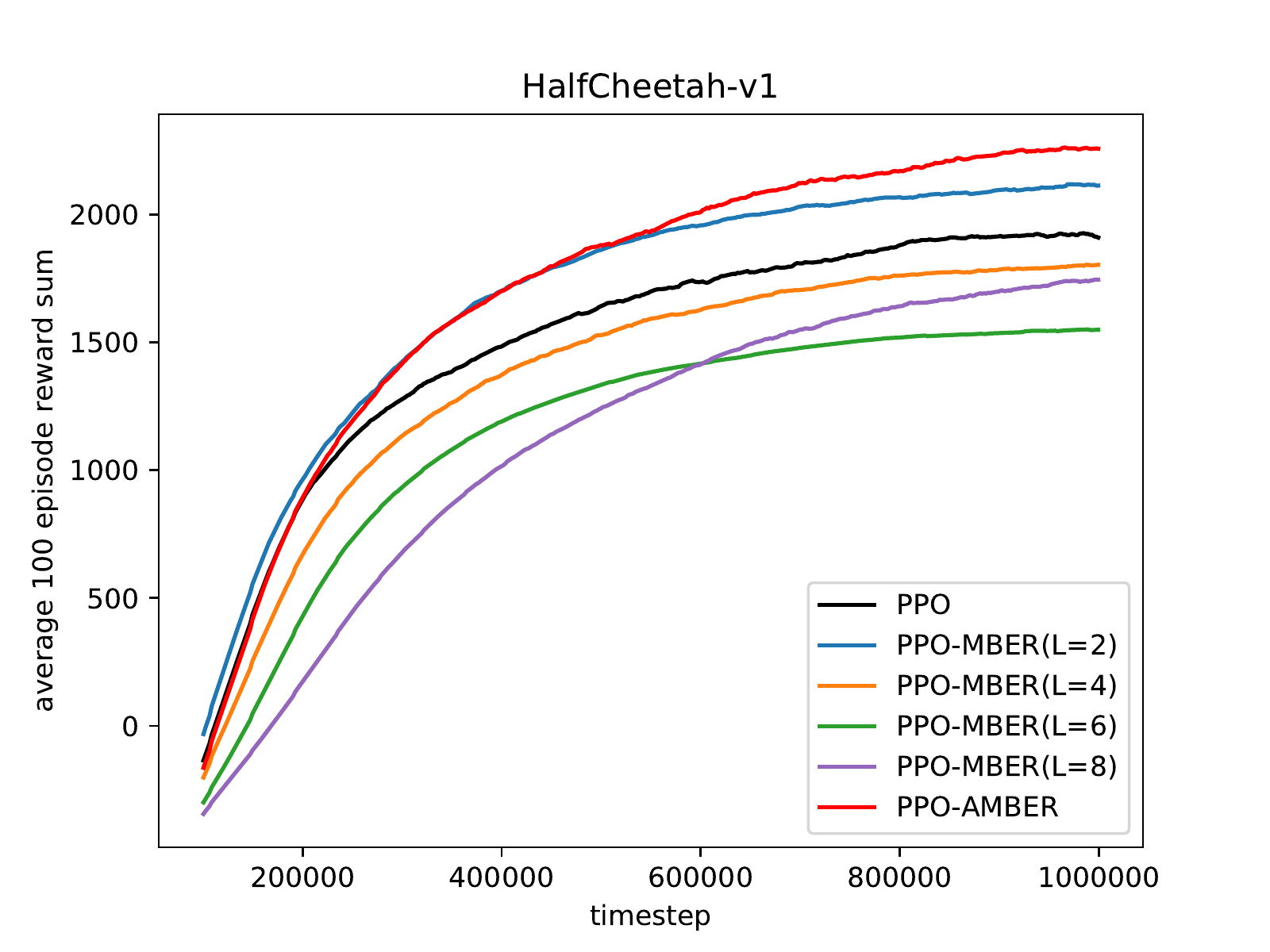}
	\includegraphics[width=0.32\textwidth]{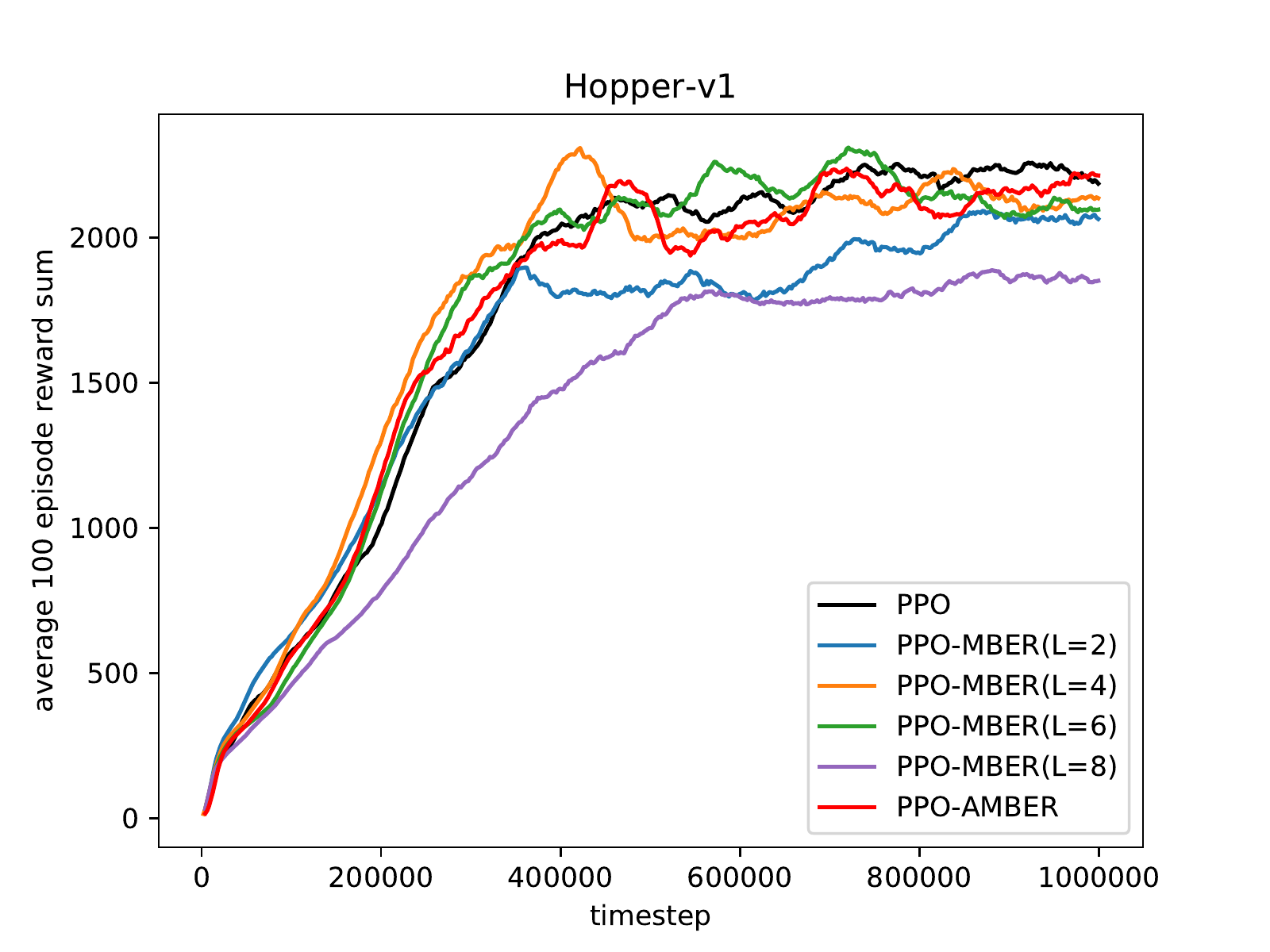}
	\includegraphics[width=0.32\textwidth]{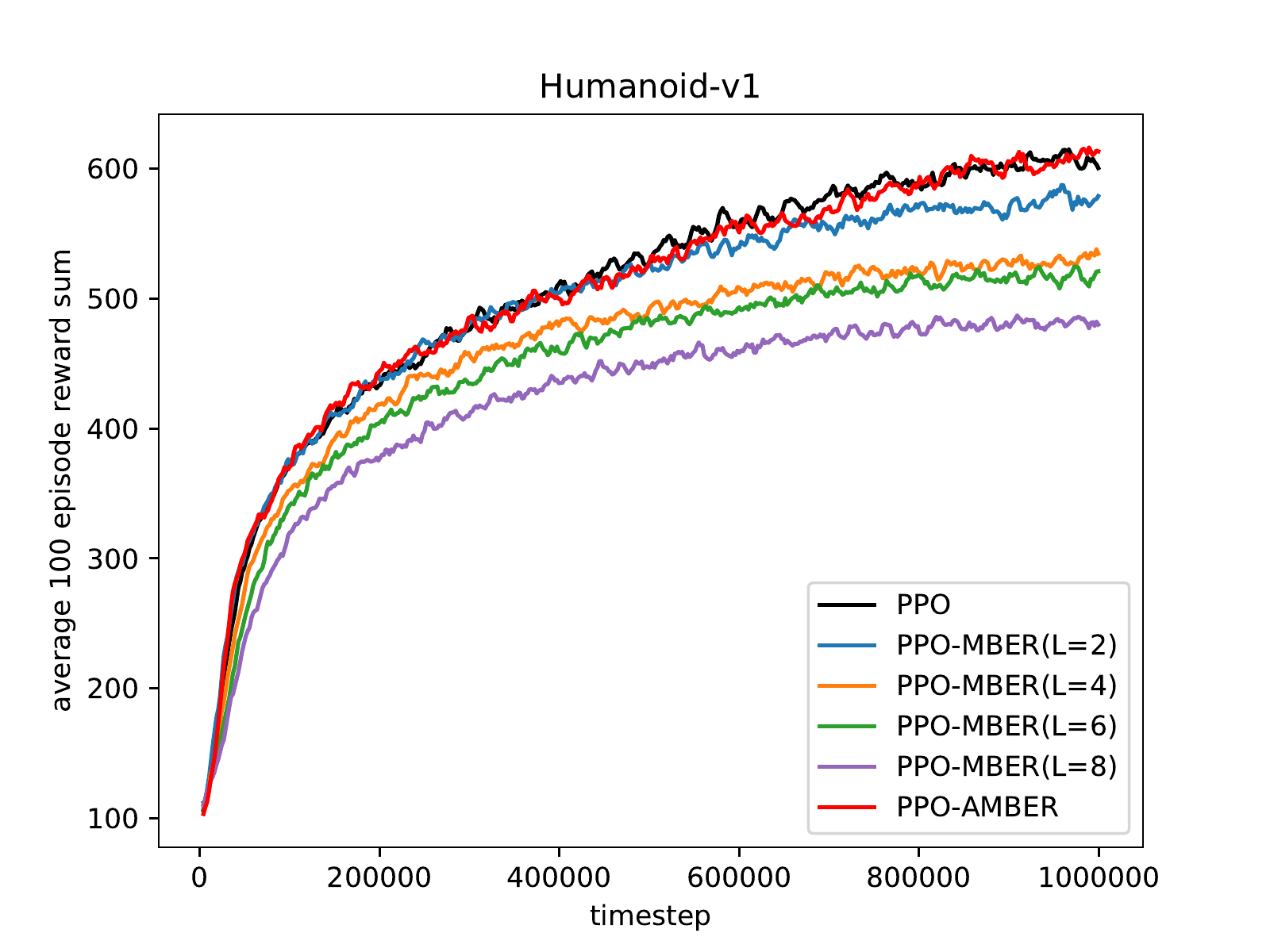}
	\includegraphics[width=0.32\textwidth]{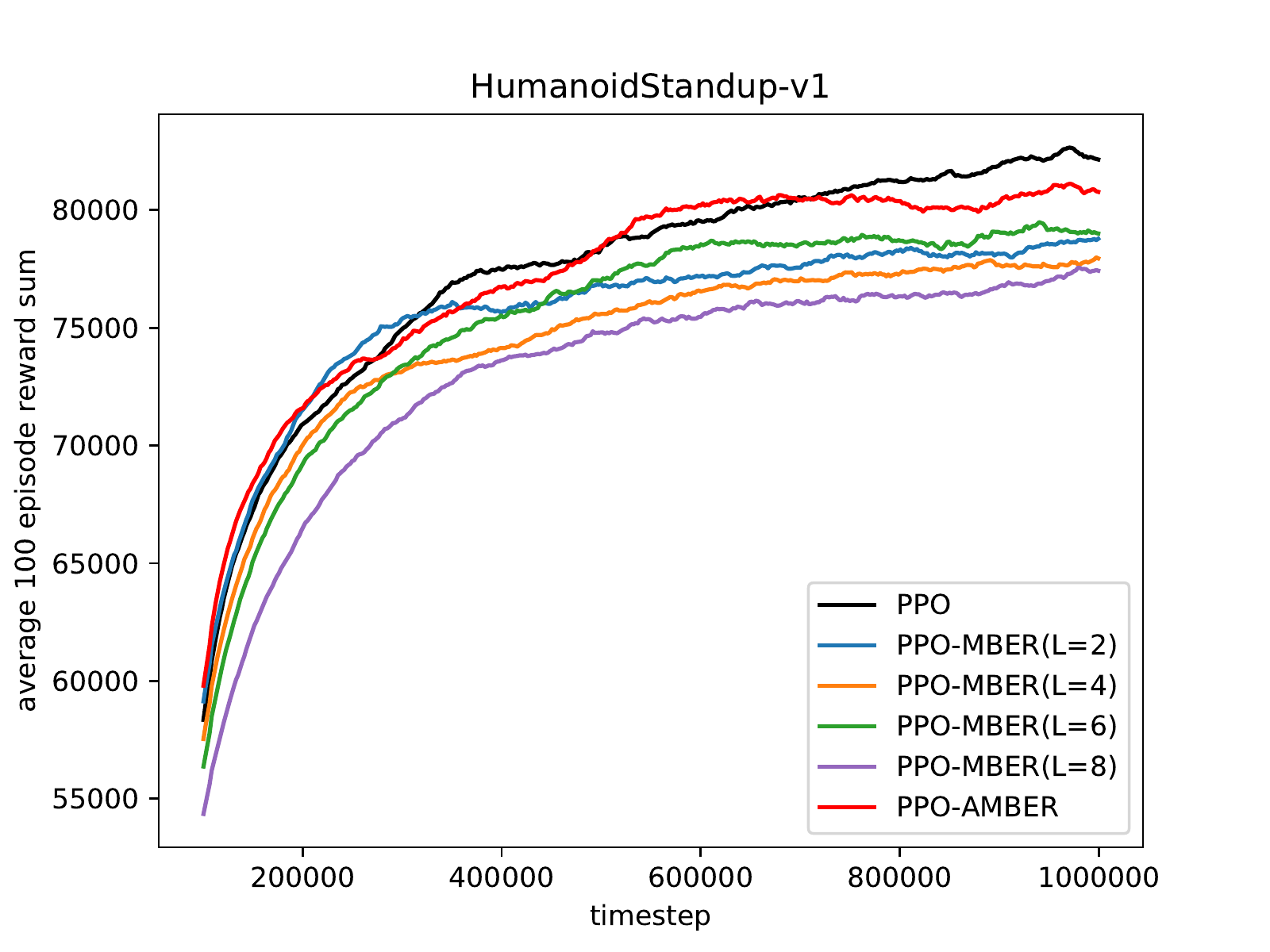}
	\includegraphics[width=0.32\textwidth]{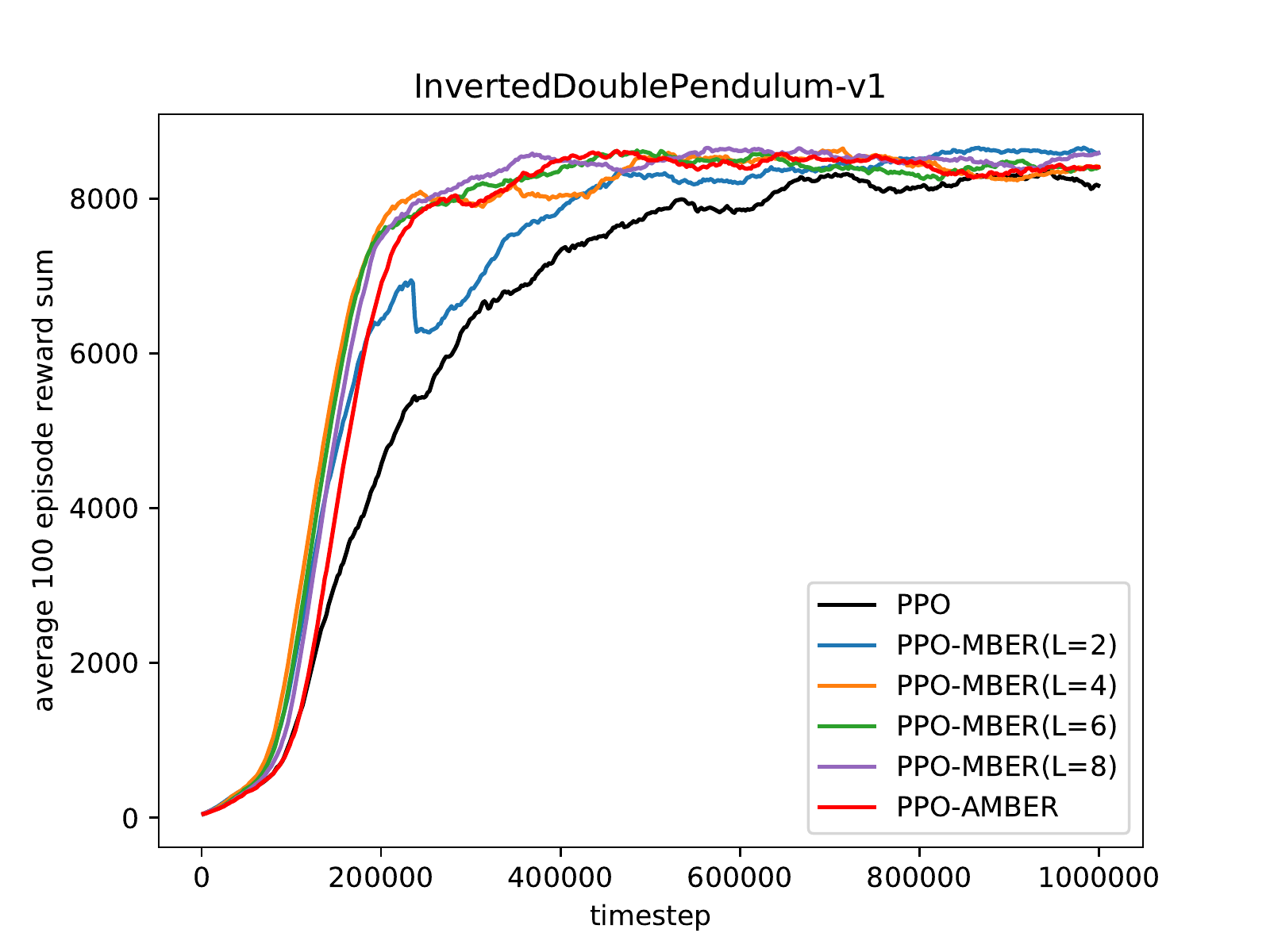}
	\includegraphics[width=0.32\textwidth]{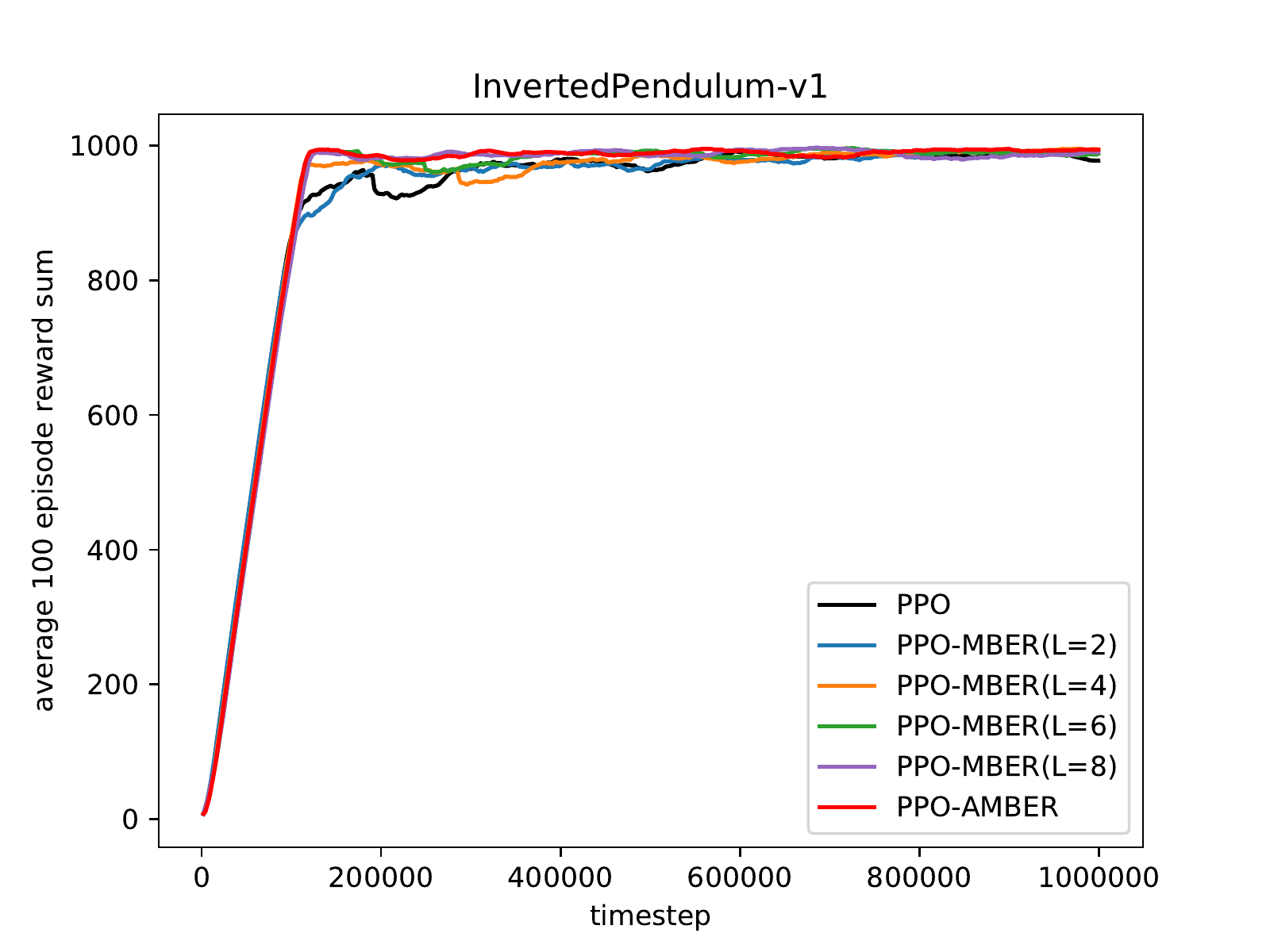}
	\includegraphics[width=0.32\textwidth]{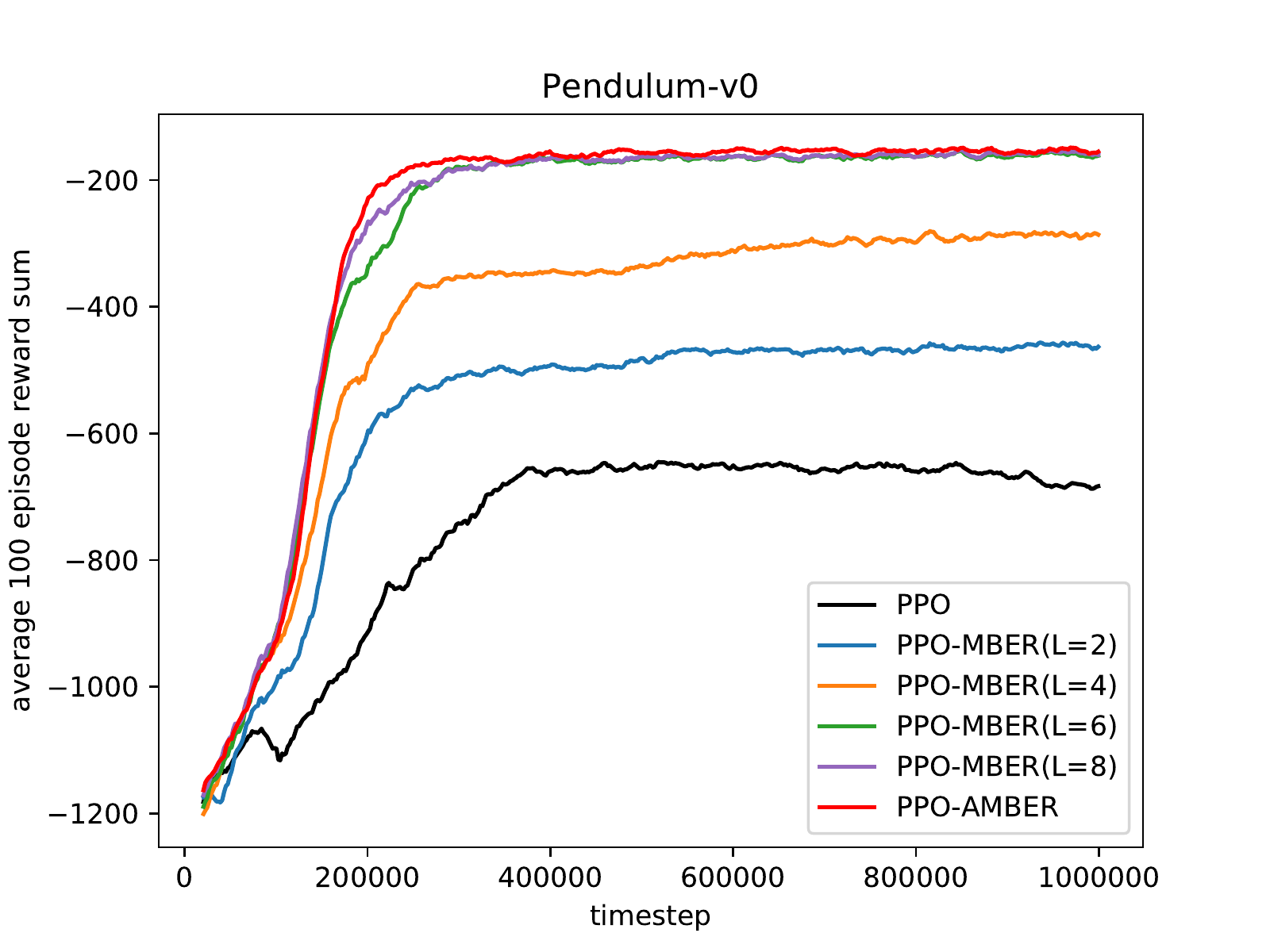}
	\includegraphics[width=0.32\textwidth]{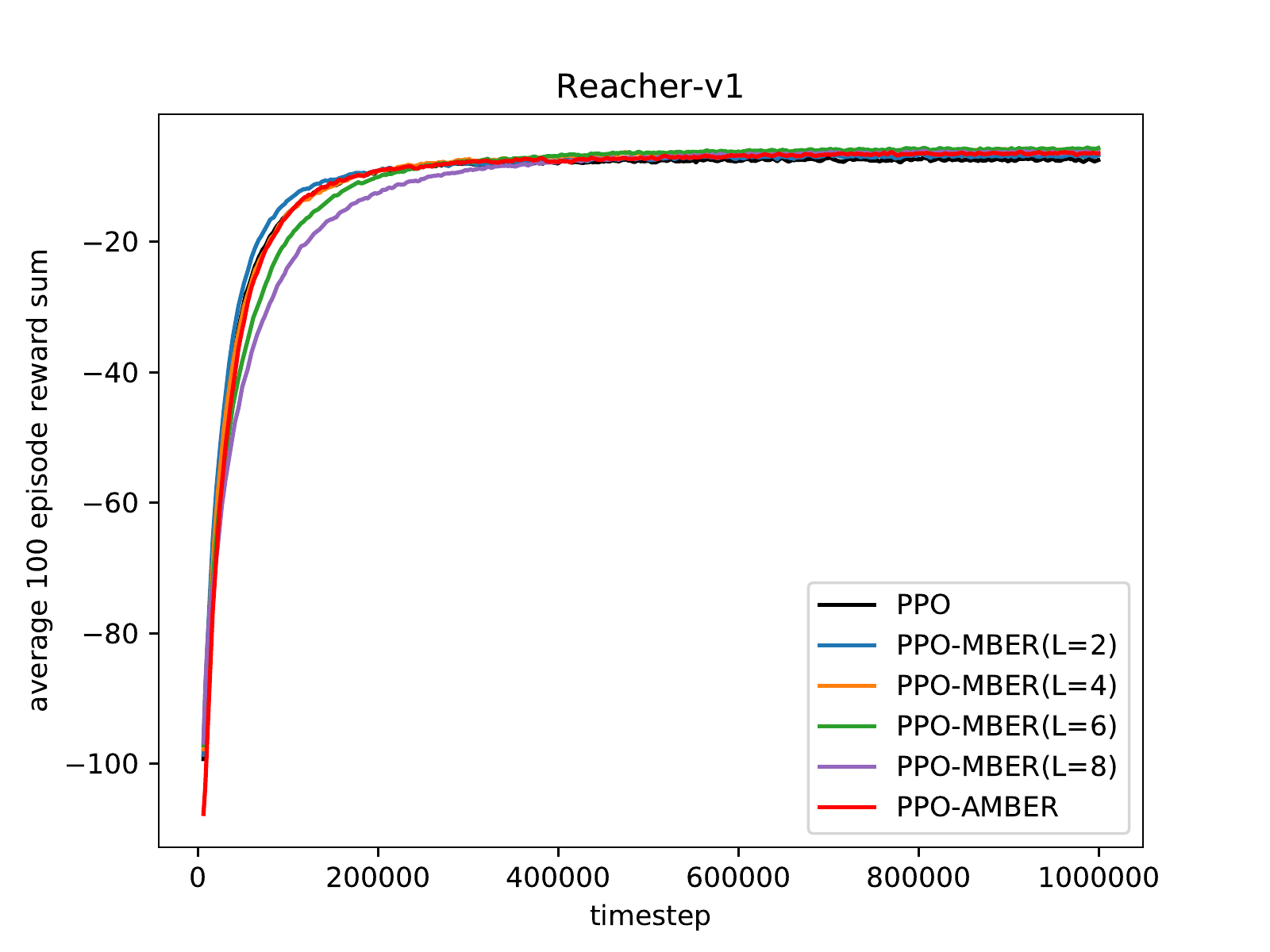}
	\includegraphics[width=0.32\textwidth]{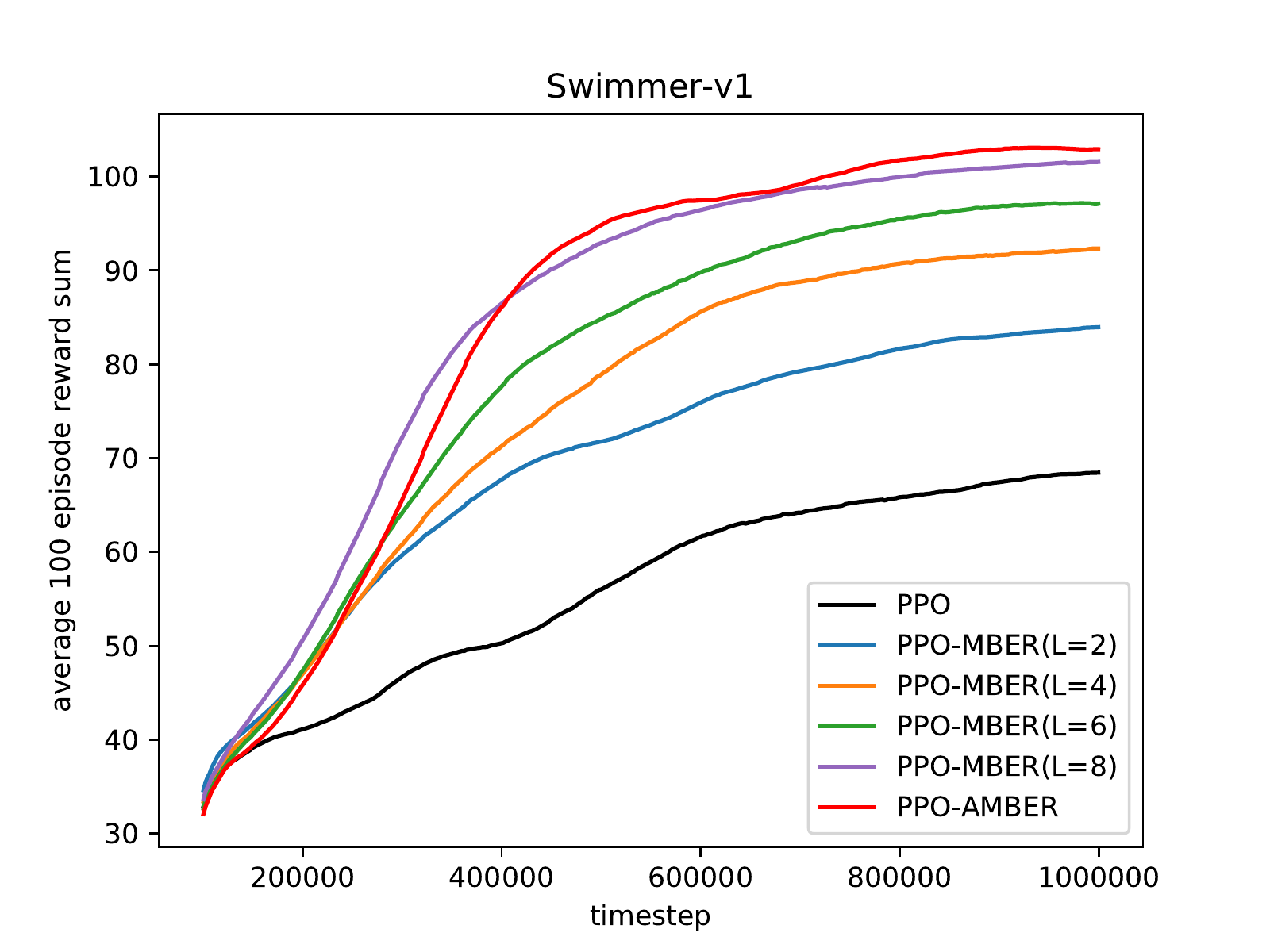}
	\includegraphics[width=0.32\textwidth]{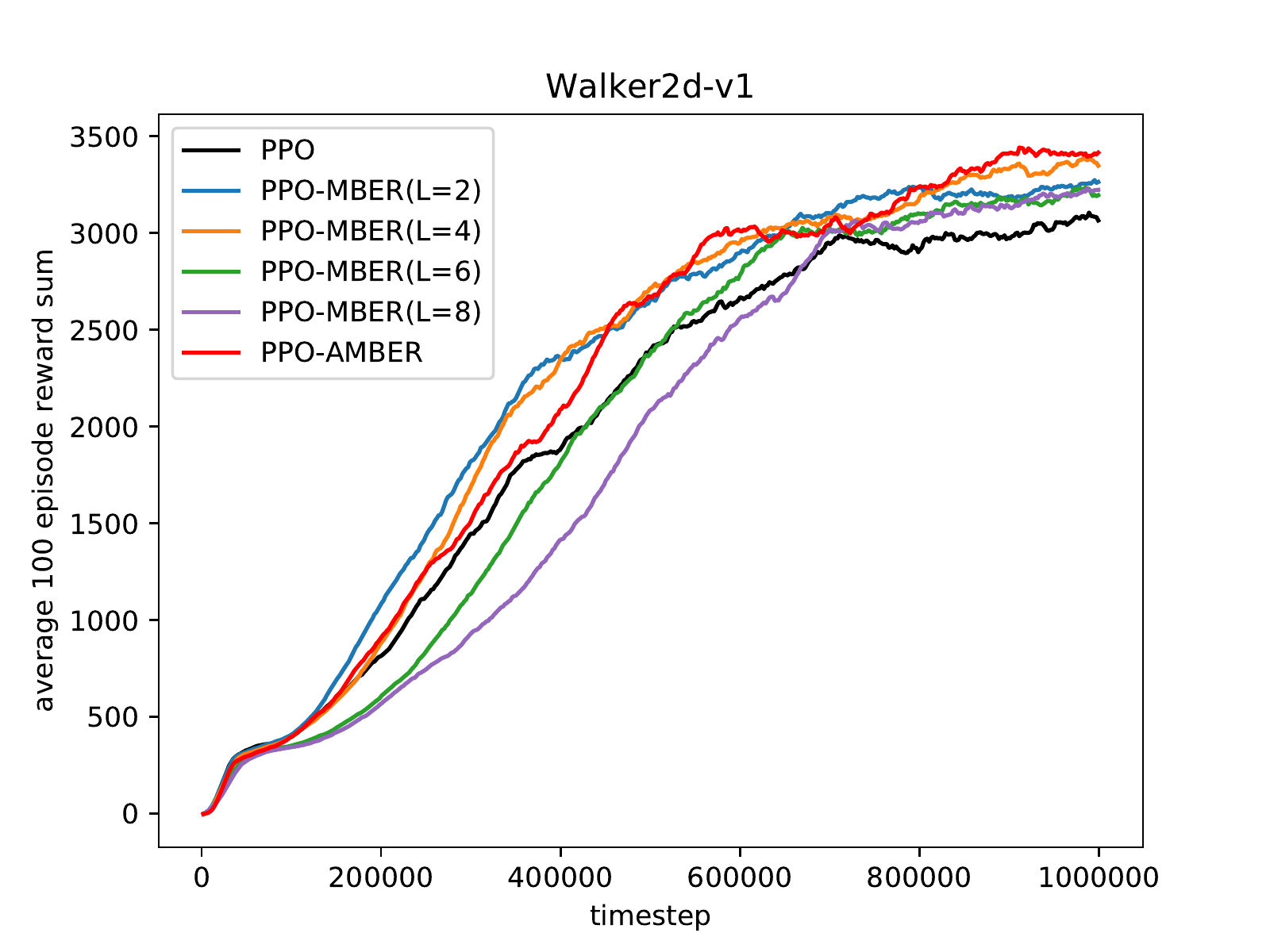}
	\caption{Performance comparison on continuous control tasks for parameter-tuned setup}
	\label{fig:NumPlotFinal}
\end{figure*}

From the ablation study in the previous subsection, we set $\epsilon = 0.4$, which is good for a wide range of taskts, and set $L=8$ as the maximum replay size  for PPO-AMBER.  PPO-AMBER shrinks the mini-batch size as $M=64\times\textrm{\# of active batches}$, as shown in Algorithm \ref{algorithm:PPO-AMBER}, and other parameters are the same as those of PPO-MBER, as shown in Table \ref{table:PPOparam}.
The batch drop factor $\epsilon_b$ is linearly annihilated from the initial value to zero as time step goes on.
 To search for optimal batch drop factor, we sweep the initial value of the batch drop factor from $0.1$ to $0.3$ and the corresponding ANS result of PPO-AMBER is provided in Table \ref{table:AdaptiveFinalANS} and \ref{table:AdaptiveFastANS}.
In addition, Fig. \ref{fig:AdaptiveBatch} shows the number of active batches of PPO-AMBER for various  batch drop factors for Pendulum, BipedalWalkerHardcore, and Humanoid tasks. It is seen from the result that $\epsilon_b=0.25$ seems appropriate.
 Tables \ref{table:LastResult} and \ref{table:AvgResult} and Fig. \ref{fig:NumPlotFinal} show the performance of PPO ($\epsilon=0.3$), PPO-MBER ($\epsilon=0.4$), and PPO-AMBER ($L=8,~\epsilon=0.4,~\epsilon_b=0.25$) on various tasks.
It is seen that PPO-AMBER with $\epsilon_b=0.25$ automatically selects almost optimal replay size  from $L=1$ to $L=8$.
So, with PPO-AMBER one need not be concerned about designing the replay memory size for the proposed ER scheme, and it significantly enhances the overall performance. We also compared the performance of PPO-AMBER with other PG methods (TRPO, ACER) in Appendix \ref{subsec:ComparePG}, and it is observed that PPO-AMBER outperforms TRPO and ACER.

\subsection{Further Discussion} \label{Discussion}

It is observed that PPO-AMBER enhances the performance of tasks with low action dimensions compared to PPO by using old sample batches, but it is hard to improve tasks with high action dimensions such as Humanoid and HumanoidStandup. This is because higher action dimensions yields  larger IS weights. Hence, we provide an additional IS analysis for those environments in Appendix \ref{subsec:ISHigh}. The analysis suggests that AMBER fits to low action dimensional tasks or sufficiently small learning rates to prevent that IS weights become too large.

\section{Conclusion} \label{sec:Conclusion}

In this paper, we have proposed a MBER scheme for PPO-style IS-based PG, which
significantly enhances the speed and  stability of convergence on various continuous control tasks (Mujoco tasks, classic control, and Box2d on OpenAI GYM) by 1) increasing the sample  efficiency without causing much bias by fixing the number of updates and reducing the IS weight, 2) reducing the sample correlation by drawing random mini-batches  with the pre-computed and stored advantages and values, and 3) dropping too old samples in the replay memory  adaptively. We have provided ablation studies on the proposed scheme, and numerical results show that the proposed method, PPO-AMBER, significantly original PPO.

\clearpage

\nocite{langley00}

\bibliography{referenceBibs_arxiv}
\bibliographystyle{plain}
\clearpage
\onecolumn

\section{Appendix} \label{sec:Appendix}

\subsection{Algorithm}
\begin{algorithm*}[!h]
	\caption{Proximal Policy Optimization with (Adaptive) Multi-Batch Experience Replay}\label{algorithm:PPO-AMBER}
	\begin{algorithmic}[1]
		\REQUIRE  $N$: batch size, $M_{\textrm{PPO}}$: mini-batch size of PPO, $M$: mini-batch size, $L$: replay length, $\beta$: step size, $\epsilon_b$: batch drop factor, adaptive=1 or 0
		\STATE{Initialize $\tilde{\theta}$ and $w$.}
		\STATE{Set $\theta_0\leftarrow\tilde{\theta}$.}
		\FOR{$i=1,2,\cdots$ ~(iteration)}
		\STATE{Collect the $i$-th trajectory $(s_{i,1},a_{i,1},r_{i,1},\cdots,s_{i,N},a_{i,N},r_{i,N})$ from $\pi_{\theta_i}$.}
		\STATE{Estimate advantage functions $\hat{A}_{i,1},\cdots,\hat{A}_{i,N}$ from the trajectory.}
		\STATE{Estimate target values $\hat{V}_{i,1},\cdots,\hat{V}_{i,N}$ from the trajectory.}
		\STATE{Store the $i$-th batch $B_i = (s_{i,n},a_{i,n},\hat{A}_{i,n},\hat{V}_{i,n},\mu_{i,n},\sigma_{i})$ at the replay memory $\mathbf{R}$ of max size $NL$.}
		\IF{adaptive}
		\FOR{$l=0,\cdots,L-1$}
		\STATE{Compute $R'_{i,l}$ of batch $B_{i-l}$ in the replay memory.}
		\IF{$R'_{i,l}>1+\epsilon_b$}
		\STATE{Set a batch $B_{i-l}$ inactive}
		\ELSE
		\STATE{Set a batch $B_{i-l}$ active}
		\ENDIF
		\ENDFOR
		\STATE{Set $M$ as $M_{\textrm{PPO}}\times\textrm{\# of active batches}$}
		\ELSE
		\STATE{Set $M$ as $M_{\textrm{PPO}}\times L$, and all batches in the replay are active.}
		\ENDIF
		\FOR{$\mathrm{epoch}=1,2,\cdots,S$}
		\FOR{$j=1,2,\cdots,N/M_{\textrm{PPO}}$}
		\STATE{Draw a mini-batch of $M$ samples uniformly from the active batches.}
		\STATE{Update the parameter as $\tilde{\theta}_{ALL}\leftarrow\tilde{\theta}_{ALL} + \beta\nabla_{\tilde{\theta}_{ALL}}\hat{L}(\tilde{\theta}_{ALL})$ from the mini-batch.}
		\ENDFOR
		\ENDFOR
		\STATE{Set $\theta_{i+1}\leftarrow\tilde{\theta}$}
		\ENDFOR
	\end{algorithmic}
\end{algorithm*}

\clearpage
\subsection{Parameter Setup}

The detailed parameter setup for PPO and PPO-MBER is given in Table \ref{table:PPOparam}. Clipping factor $\epsilon$ of PPO and PPO-MBER varies from $0.2$ to $0.7$. We set $\epsilon$ of PPO-AMBER as $\epsilon=0.4$. The batch drop factor $\epsilon_b$ is varied from $0.1$ to $0.3$. The mini-batch size of PPO-AMBER adaptively changes as $M=64\times \#$ of active batches.

\begin{table*}[!h]\footnotesize
	\caption{Hyperparameters of PPO, PPO-MBER, and PPO-AMBER}
	\label{table:PPOparam}
	\centering
	\begin{tabular}{lC{1.9cm}C{1.9cm}C{1.9cm}}
		\toprule
		Hyperparameter & PPO & PPO-MBER & PPO-AMBER\\
		\cmidrule{1-4}
		Initial batch drop factor ($\epsilon_b$) & $\cdot$ & $\cdot$ & variable \\
		Replay length ($L$) & $\cdot$ & $2, 4, 6, 8$ & $8$\\
		mini-batch size ($M$) & $64$ & $64L$ & adaptive \\
		Initial clipping factor ($\epsilon$) & variable & variable & 0.4\\
		
		Horizon ($N$) & $2048$ &$2048$ & {$2048$} \\
		Initial Adam step size ($\beta$) & {$3\cdot10^{-4}$} & {$3\cdot10^{-4}$} & {$3\cdot10^{-4}$} \\
		Epochs ($S$) & {$10$} & $10$ & $10$ \\
		Discount factor ($\gamma$) & {$0.99$} & $0.99$ & $0.99$ \\
		TD parameter ($\lambda$) & {$0.95$} &  {$0.95$}  & {$0.95$} \\
		\bottomrule
	\end{tabular}
\end{table*}

\subsection{Simulation Results}

We provide the simulation results of parameter-tuned PPO, PPO-MBER, and PPO-AMBER on tasks  BipedalWalker, BipedalWalkerHardcore, HalfCheetah, Hopper, Humanoid, HumanoidStandup, InvertedDoublePendulum, InvertedPendulum, Pendulum, Reacher,  Swimmer, and Walker2d.

\begin{table*}[!h]\footnotesize
	\caption{Average return of final 100 episodes and the corresponding final ANS of parameter-tuned PPO, PPO-MBER and PPO-AMBER} \label{table:LastResult}
	\centering
	\resizebox{\textwidth}{!}{
	\begin{tabular}{lC{2.1cm}C{2.1cm}C{2.1cm}C{2.1cm}C{2.1cm}C{2.1cm}}
		\toprule
		& PPO & PPO-MBER $(L=2)$ & PPO-MBER $(L=4)$ & PPO-MBER $(L=6)$ & PPO-MBER $(L=8)$ & PPO-AMBER\\
		\cmidrule{1-7}
		BipedalWalker & $236\pm26$ & $\mathbf{276\pm16}$ & $275\pm19$ & $261\pm13$ & $225\pm75$ & $265\pm16$ \\
		BipedalWalkerHardcore & $-93.5\pm10.8$ & $-77.9\pm20.8$ & $-80.7\pm18.4$ & $\mathbf{-73.2\pm18.8}$ & $-78.1\pm15.8$ & $-73.5\pm10.7$ \\
		HalfCheetah & $1910\pm778$ & $2113\pm874$ & $1803\pm611$ & $1549\pm562$ & $1745\pm529$ & $\mathbf{2258\pm1039}$ \\
		Hopper & $2185\pm371$ & $2063\pm286$ & $2135\pm342$ & $2095\pm316$ & $1850\pm613$ & $\mathbf{2213\pm295}$ \\
		Humanoid & $600\pm43$ & $578\pm41$ & $534\pm34$ & $521\pm24$ & $480\pm14$ & $\mathbf{613\pm67}$ \\
		HumanoidStandup & $\mathbf{82149\pm3681}$ & $78782\pm5620$ & $77948\pm3852$ & $79004\pm4377$ & $77425\pm5461$ & $80774\pm3518$ \\
		InvertedDoublePendulum & $8167\pm630$ & $\mathbf{8588\pm271}$ & $8407\pm305$ & $8402\pm209$ & $8587\pm208$ & $8406\pm363$ \\
		InvertedPendulum & $977\pm20$ & $992\pm12$ & $\mathbf{993\pm6}$ & $987\pm13$ & $989\pm11$ & $993\pm9$ \\
		Pendulum & $-683\pm494$ & $-463\pm391$ & $-286\pm306$ & $-161\pm7$ & $-160\pm8$ & $\mathbf{-155\pm12}$ \\
		Reacher & $-7.5\pm2.1$ & $-6.7\pm2.3$ & $-5.9\pm1.0$ & $\mathbf{-5.6\pm0.9}$ & $-6.3\pm0.8$ & $-6.5\pm1.2$ \\
		Swimmer & $68.4\pm19.0$ & $83.9\pm21.5$ & $92.3\pm21.5$ & $97.1\pm24.5$ & $101.6\pm26.2$ & $\mathbf{102.9\pm26.4}$ \\
		Walker2d & $3065\pm532$ & $3264\pm577$ & $3348\pm498$ & $3196\pm429$ & $3223\pm445$ & $\mathbf{3415\pm416}$ \\
		\midrule
		Final ANS & $0.176$ & $0.562$ & $0.533$ & $0.495$ & $0.279$ & $\mathbf{0.939}$ \\
		\bottomrule
	\end{tabular}
	}
	\vspace{1em}\\
	\caption{Average return of all episodes and the corresponding speed ANS of parameter-tuned PPO, PPO-MBER and PPO-AMBER} \label{table:AvgResult}
	\resizebox{\textwidth}{!}{
	\begin{tabular}{lC{2.1cm}C{2.1cm}C{2.1cm}C{2.1cm}C{2.1cm}C{2.1cm}}
		\toprule
		& PPO & PPO-MBER ($L=2$) & PPO-MBER ($L=4$) & PPO-MBER ($L=6$) & PPO-MBER ($L=8$) & PPO-AMBER \\
		\cmidrule{1-7}
		BipedalWalker & $188\pm29$ & $\mathbf{219\pm25}$ & $204\pm29$ & $198\pm25$ & $158\pm64$ & $213\pm19$ \\
		BipedalWalkerHardcore & $-107.2\pm5.3$ & $-97.7\pm14.6$ & $-96.2\pm13.6$ & $-94.0\pm12.6$ & $-94.9\pm9.3$ & $\mathbf{-90.5\pm5.7}$ \\
		HalfCheetah & $1511\pm583$ & $1693\pm630$ & $1383\pm365$ & $1167\pm367$ & $1127\pm234$ & $\mathbf{1728\pm709}$ \\
		Hopper & $1736\pm241$ & $1604\pm112$ & $\mathbf{1761\pm250}$ & $1750\pm198$ & $1390\pm493$ & $1719\pm99$ \\
		Humanoid & $\mathbf{506\pm26}$ & $495\pm23$ & $462\pm19$ & $449\pm11$ & $421\pm11$ & $504\pm33$ \\
		HumanoidStandup & $\mathbf{77484\pm3032}$ & $75729\pm4806$ & $74568\pm3372$ & $75486\pm3311$ & $73134\pm4322$ & $77239\pm2857$ \\
		InvertedDoublePendulum & $6399\pm442$ & $6961\pm296$ & $7265\pm131$ & $7256\pm153$ & $\mathbf{7280\pm115}$ & $7104\pm99$ \\
		InvertedPendulum & $918\pm12$ & $918\pm9$ & $922\pm8$ & $926\pm7$ & $927\pm3$ & $\mathbf{930\pm4}$ \\
		Pendulum & $-751\pm372$ & $-567\pm333$ & $-421\pm276$ & $-288\pm38$ & $-279\pm31$ & $\mathbf{-272\pm10}$ \\
		Reacher & $-10.9\pm1.6$ & $-10.2\pm2.0$ & $\mathbf{-10.0\pm0.9}$ & $-10.7\pm0.8$ & $-12.0\pm0.1$ &  $-10.7\pm1.0$ \\
		Swimmer & $56.2\pm9.4$ & $69.7\pm16.4$ & $75.5\pm14.9$ & $79.5\pm19.6$ & $\mathbf{85.3\pm21.8}$ & $84.8\pm20.8$ \\
		Walker2d & $2015\pm400$ & $\mathbf{2248\pm403}$ & $2230\pm424$ & $2006\pm299$ & $1868\pm313$ & $2208\pm320$ \\
		\midrule
		Speed ANS & $0.274$ & $0.570$ & $0.556$ & $0.518$ & $0.076$ & $\mathbf{0.979}$ \\
		\bottomrule
	\end{tabular}
	}
\end{table*}

\begin{table*}[!h]\footnotesize
	\centering
	\caption{Final ANS for PPO and PPO-MBER}\label{table:FinalANS}
	\begin{tabular}{cC{1.85cm}C{1.85cm}C{1.85cm}C{1.85cm}C{1.85cm}}
		\toprule
		Clipping factor ($\epsilon$) & PPO & PPO-MBER ($L=2$) & PPO-MBER ($L=4$) & PPO-MBER ($L=6$) & PPO-MBER ($L=8$) \\
		\cmidrule{1-6}
		$0.2$ & $-0.097$ & $0.475$ & $0.211$ & $-0.070$ & $-0.597$ \\
		$0.3$ & $\mathbf{0.176}$ & $0.527$ & $0.484$ & $0.192$ & $-0.130$ \\
		$0.4$ & $-0.011$ & $\mathbf{0.562}$ & $\mathbf{0.533}$ & $\mathbf{0.495}$ & $\mathbf{0.279}$ \\
		$0.5$ & $-0.678$ & $0.236$ & $0.116$ & $0.148$ & $0.110$ \\
		$0.6$ & $-1.018$ & $-0.296$ & $0.071$ & $-0.110$ & $-0.068$ \\
		$0.7$ & $-1.786$ & $-0.296$ & $-0.638$ & $-0.843$ & $-0.519$ \\
		\bottomrule
	\end{tabular}
	\vspace{1em}\\
	\centering	
	\caption{Final ANS for PPO-AMBER}\label{table:AdaptiveFinalANS}
	\begin{tabular}{cC{6cm}}
		\toprule
		Batch drop factor ($\epsilon_b$) & PPO-AMBER ($L=8,~\epsilon=0.4$)\\
		\cmidrule{1-2}
		$0.10$ & $0.196$ \\
		$0.15$ & $0.193$ \\
		$0.20$ & $0.583$ \\
		$0.25$ & $\mathbf{0.939}$ \\
		$0.30$ & $0.631$ \\
		\bottomrule
	\end{tabular}
	\vspace{1em}\\
	\centering	
	\caption{Speed ANS for PPO and PPO-MBER}\label{table:FastANS}
	\begin{tabular}{cC{1.85cm}C{1.85cm}C{1.85cm}C{1.85cm}C{1.85cm}}
		\toprule
		Clipping factor ($\epsilon$) & PPO & PPO-MBER ($L=2$) & PPO-MBER ($L=4$) & PPO-MBER ($L=6$) & PPO-MBER ($L=8$) \\
		\cmidrule{1-6}
		$0.2$ & $-0.100$ & $0.331$ & $-0.190$ & $-0.660$ & $-1.358$ \\
		$0.3$ & $\mathbf{0.274}$ & $0.508$ & $0.484$ & $0.036$ & $-0.372$ \\
		$0.4$ & $-0.002$ & $\mathbf{0.570}$ & $\mathbf{0.556}$ & $\mathbf{0.518}$ & $0.076$ \\
		$0.5$ & $-0.495$ & $0.344$ & $0.332$ & $0.309$ & $\mathbf{0.116}$ \\
		$0.6$ & $-0.949$ & $-0.188$ & $0.286$ & $0.078$ & $-0.032$ \\
		$0.7$ & $-1.580$ & $-0.311$ & $-0.439$ & $-0.536$ & $-0.334$ \\
		\bottomrule
	\end{tabular}	
	\vspace{1em}\\
	\caption{Speed ANS for PPO-AMBER}\label{table:AdaptiveFastANS}
	\begin{tabular}{cC{6cm}}
		\toprule
		Batch drop factor ($\epsilon_b$) & PPO-AMBER ($L=8,~\epsilon=0.4$)\\
		\cmidrule{1-2}
		$0.10$ & $0.083$ \\
		$0.15$ & $0.308$ \\
		$0.20$ & $0.688$ \\
		$0.25$ & $\mathbf{0.979}$ \\
		$0.30$ & $0.671$ \\
		\bottomrule
	\end{tabular}	
\end{table*}

\clearpage
\subsection{Performance Comparison with Other IS-based PG Methods}\label{subsec:ComparePG}

\begin{figure*}[!h]
	\centering
	\includegraphics[width=0.32\textwidth]{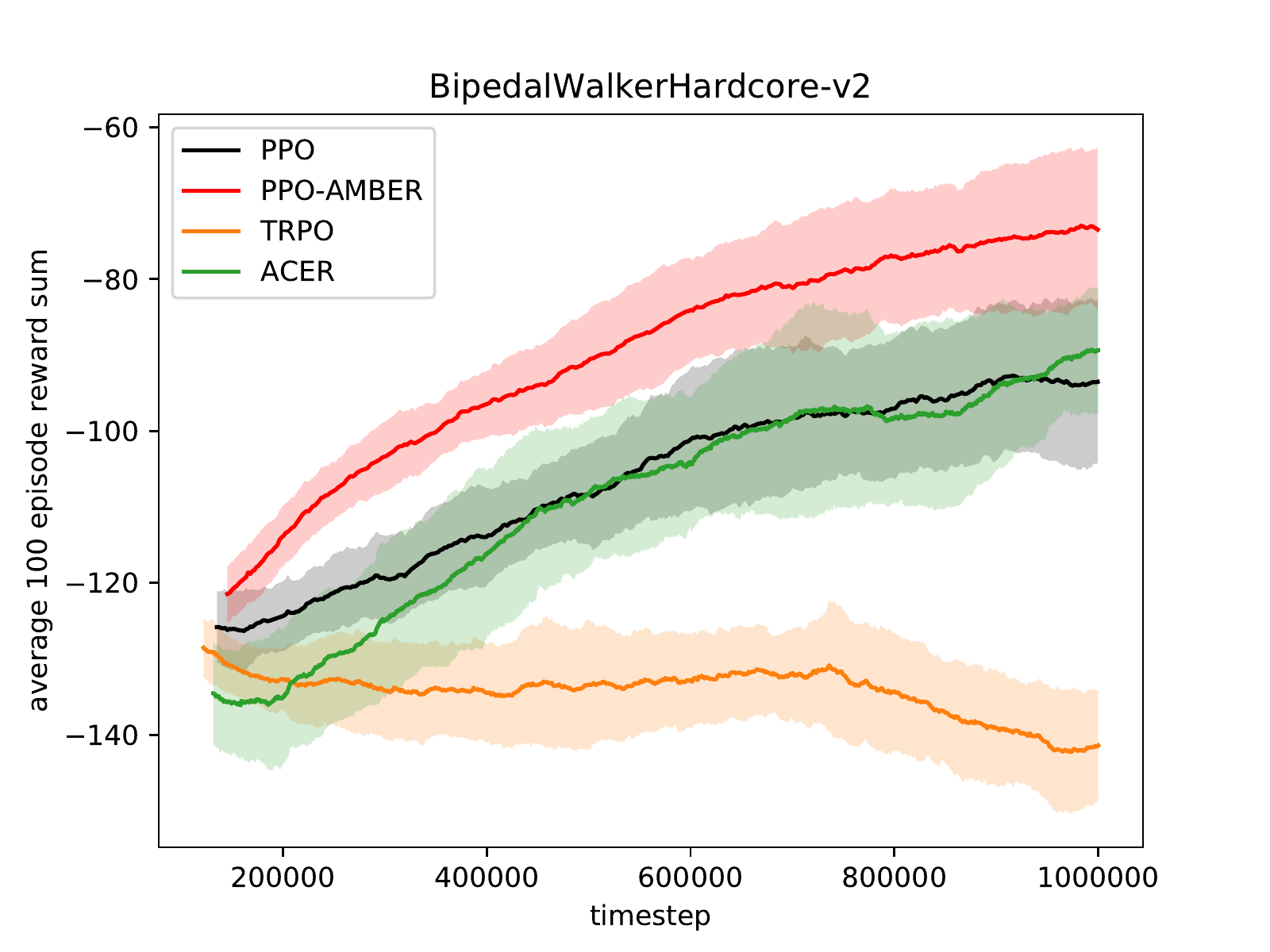}
	\includegraphics[width=0.32\textwidth]{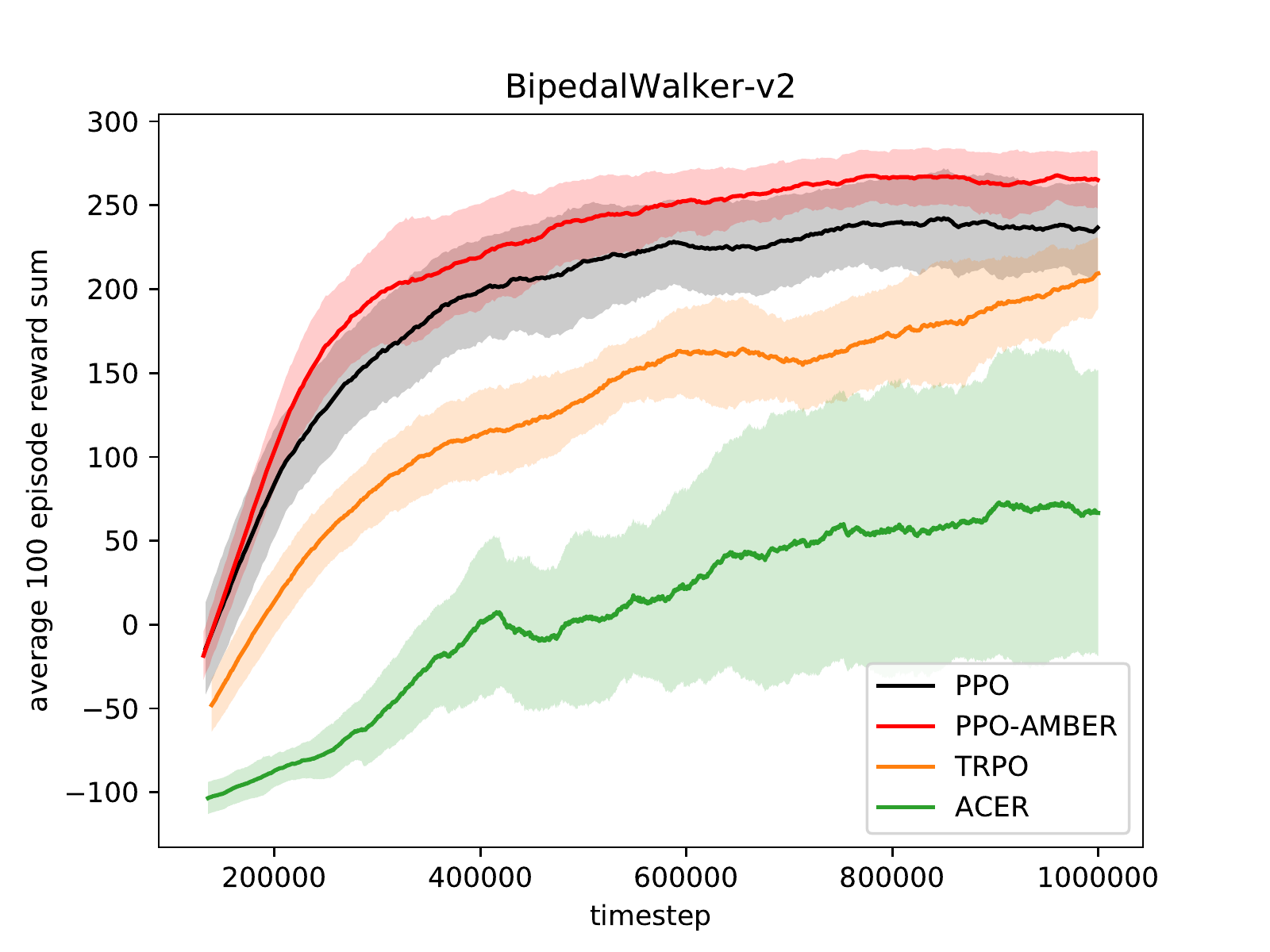}
	\includegraphics[width=0.32\textwidth]{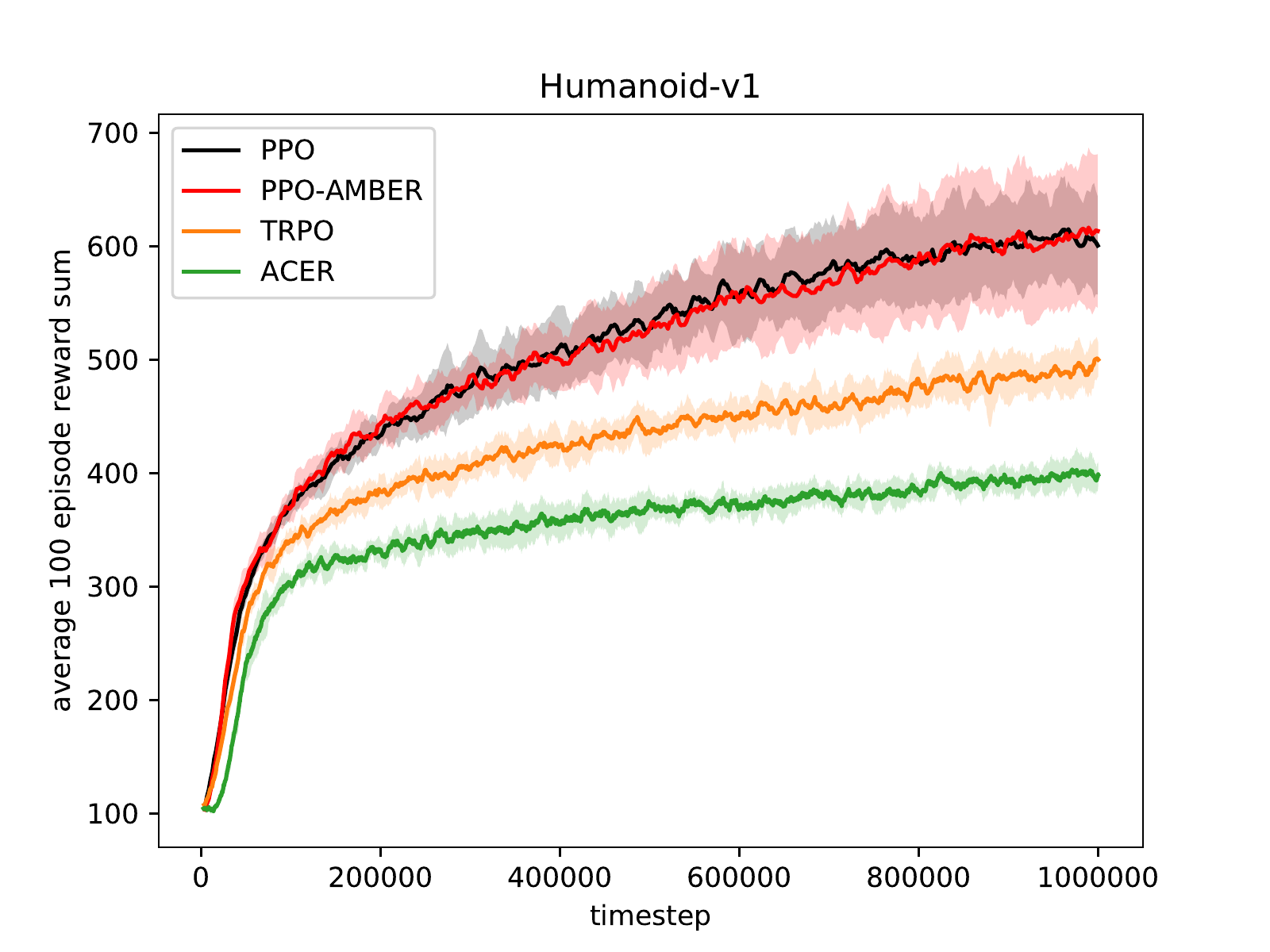}
	\includegraphics[width=0.32\textwidth]{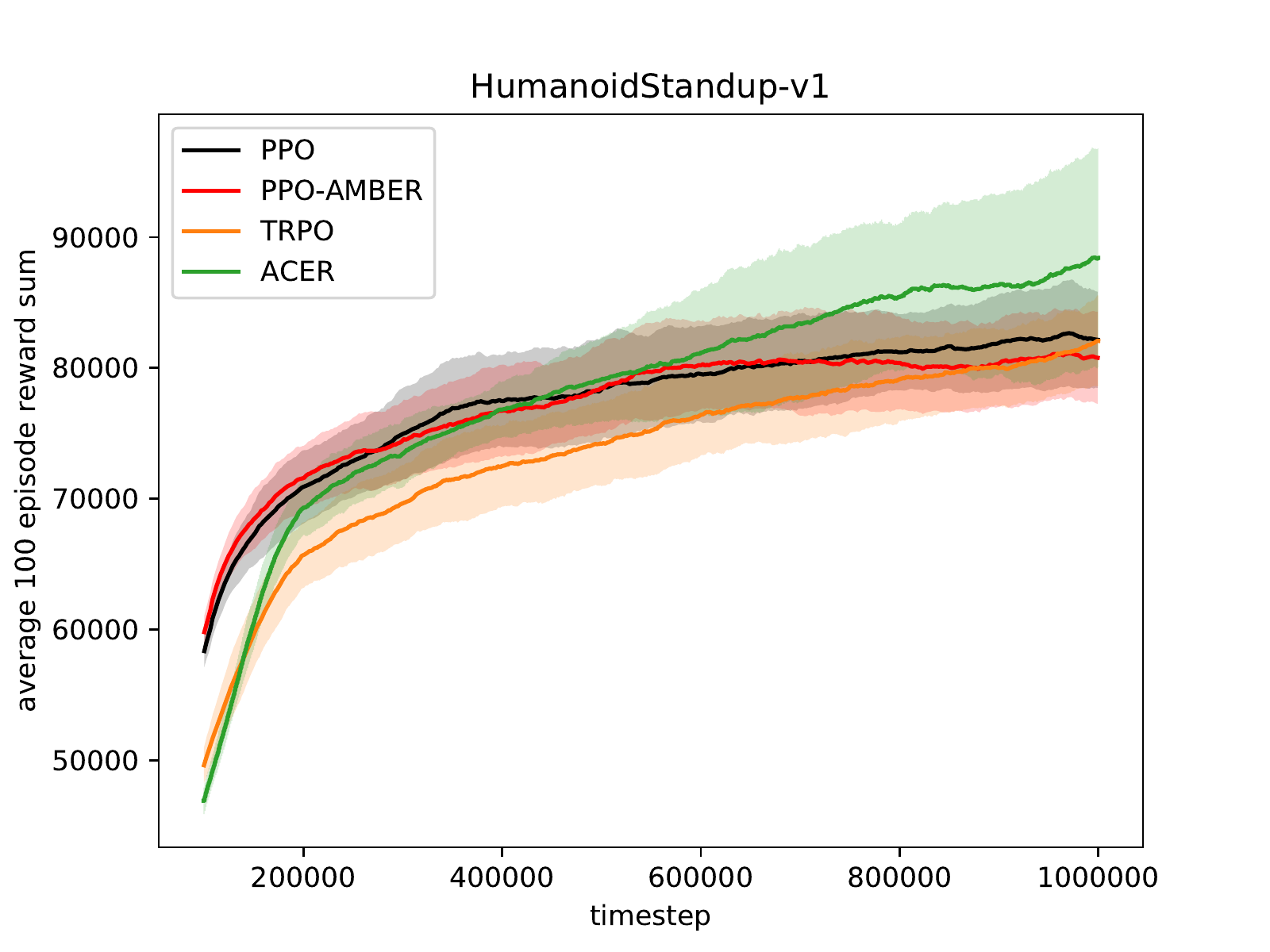}
	\includegraphics[width=0.32\textwidth]{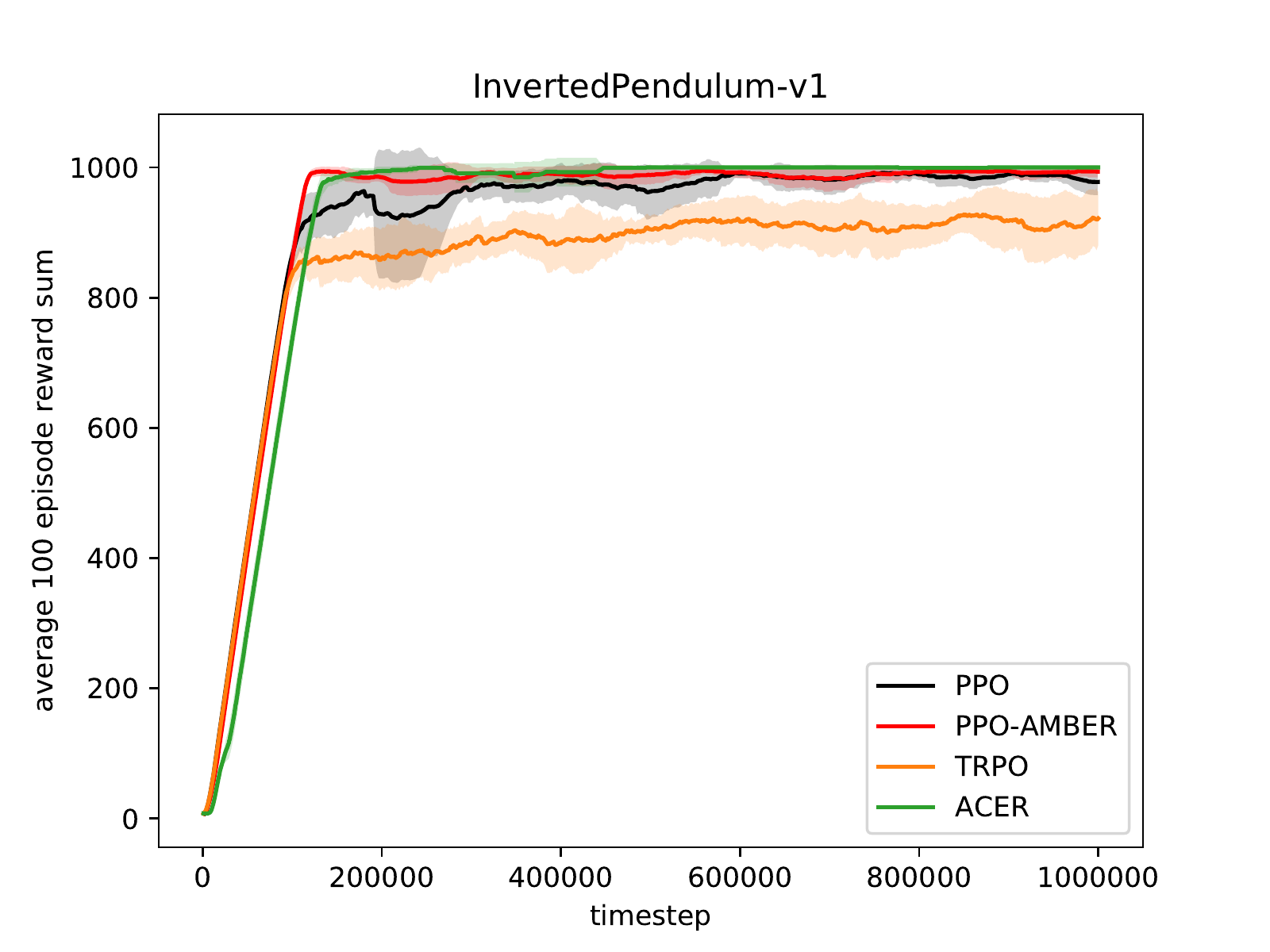}
	\includegraphics[width=0.32\textwidth]{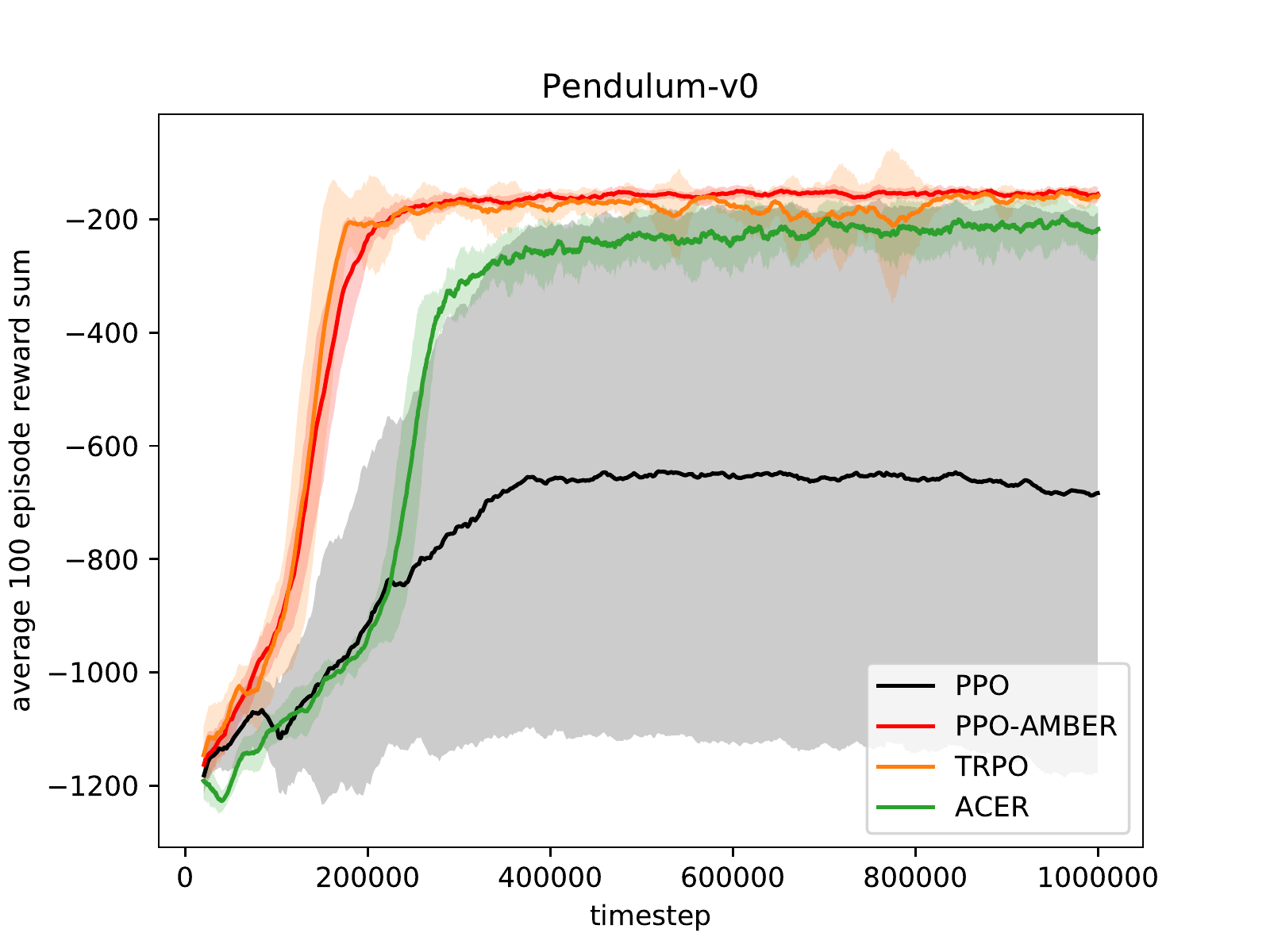}
	\caption{Performance comparison with other IS-based PG methods}
	\label{fig:NumPlotComparison}
\end{figure*}

Since the paper considers the performance improvement for IS-based PG methods based on ER, we compared proposed PPO-AMBER with other IS-based PG methods: TRPO and ACER. We used the single-path TRPO of OpenAI baselines and our own ACER which is a modified version from the discrete ACER in OpenAI baselines. The policy networks of both algorithms were a Gaussian policy which was the same as that of PPO. For TRPO, the batch size was 1024 and  the KL step size was 0.01. Note that the batch size and the performance of TRPO  fit to  1M time-step simulation as seen in the result comparison in \cite{schulman2017proximal}. For ACER, the stochastic dueling network with 2 hidden layers of size $64$ and $n=5$  \cite{wang2016sample} was used for the value network, and we used the fixed learning rate $7\cdot10^{-4}$ and the KL step size $0.1$. Other parameters of ACER were the same as \cite{wang2016sample}. The result is given in Fig. \ref{fig:NumPlotComparison}. It is seen that PPO-AMBER outperforms TRPO and ACER.

\subsection{IS Analysis on High Action-Dimensional Tasks}\label{subsec:ISHigh}

\begin{figure}[!h]\label{fig:LR}
	\centering
	\includegraphics[width=0.4\textwidth]{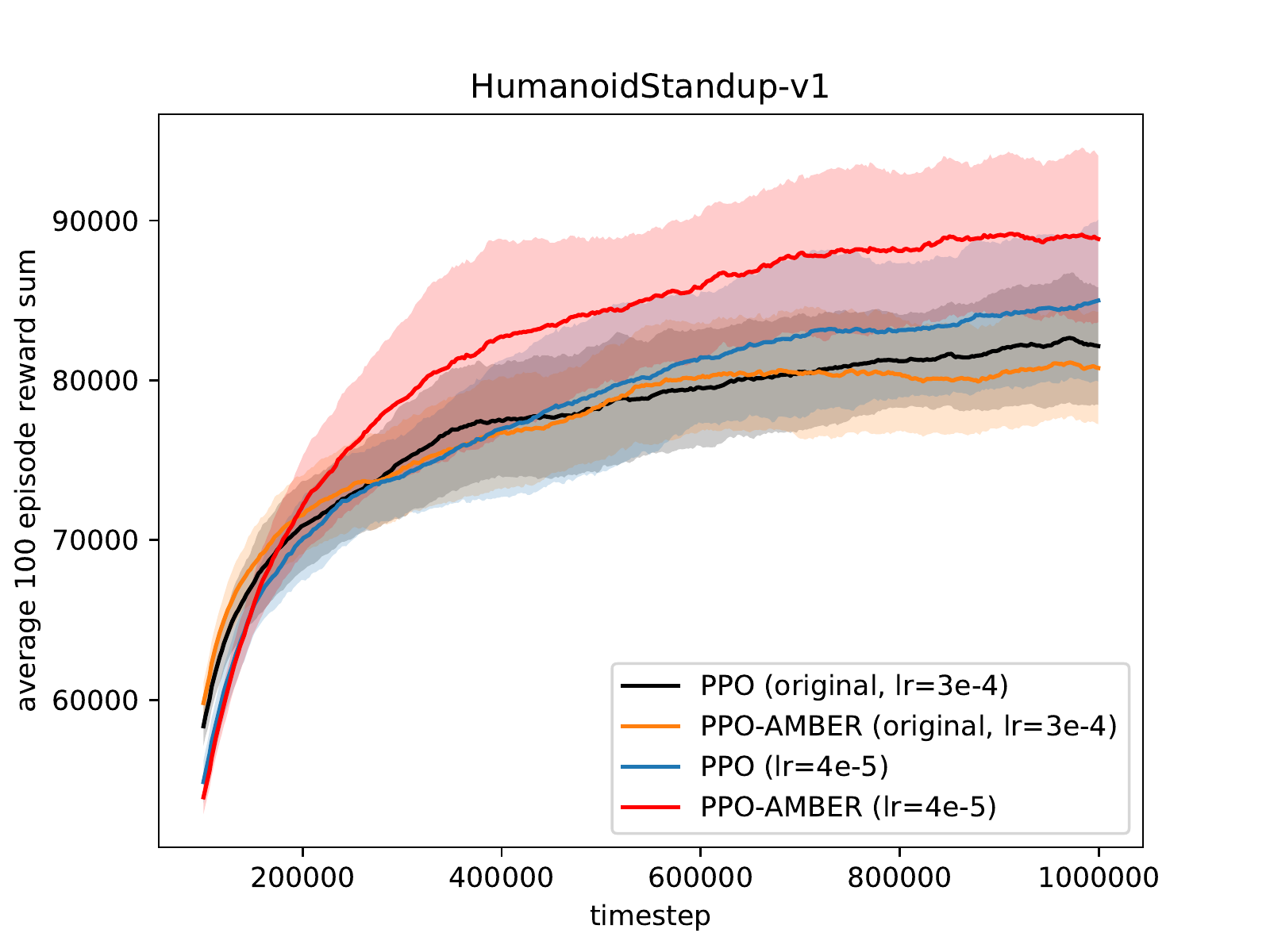}
	\includegraphics[width=0.4\textwidth]{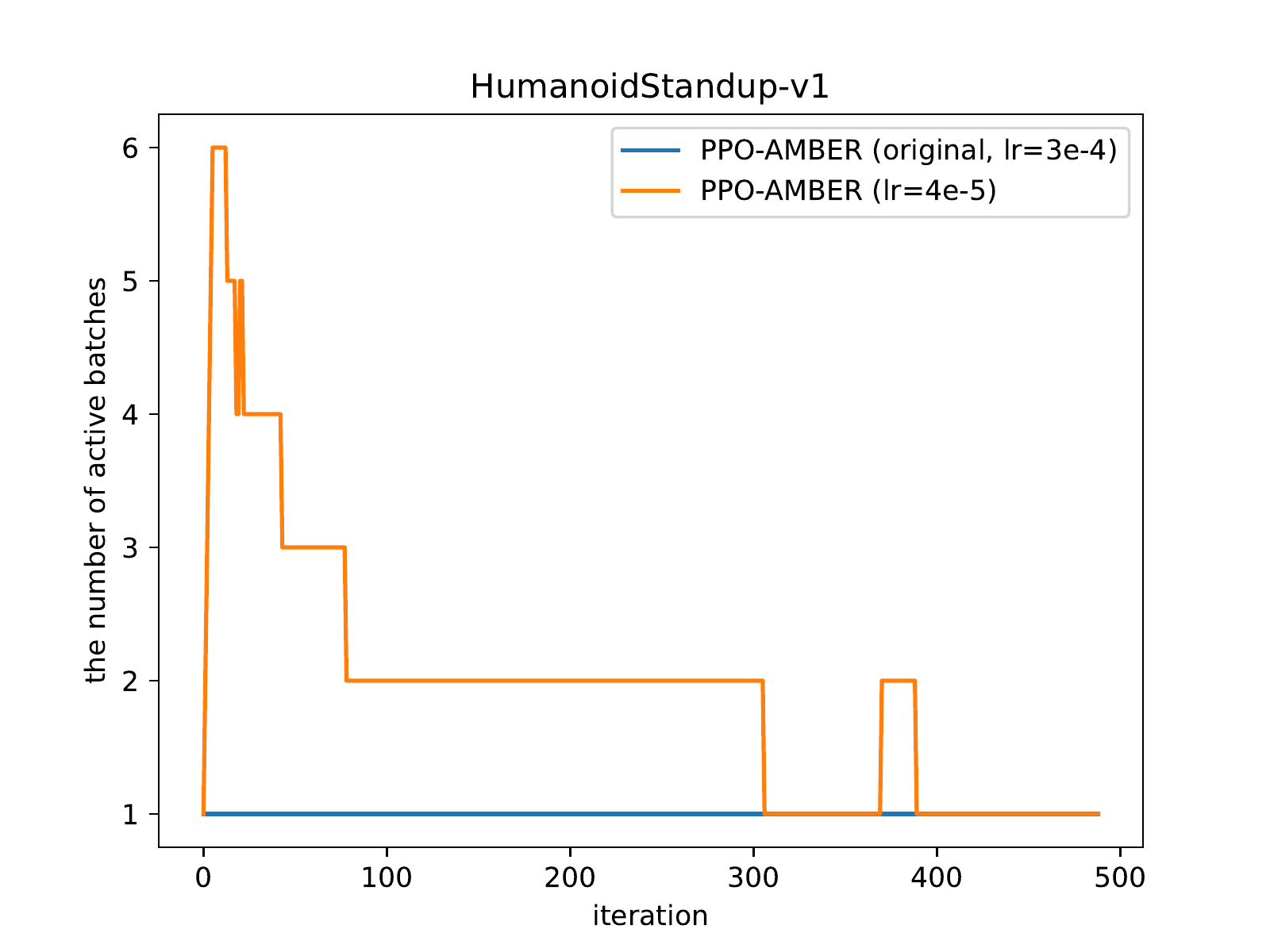}
	\includegraphics[width=0.4\textwidth]{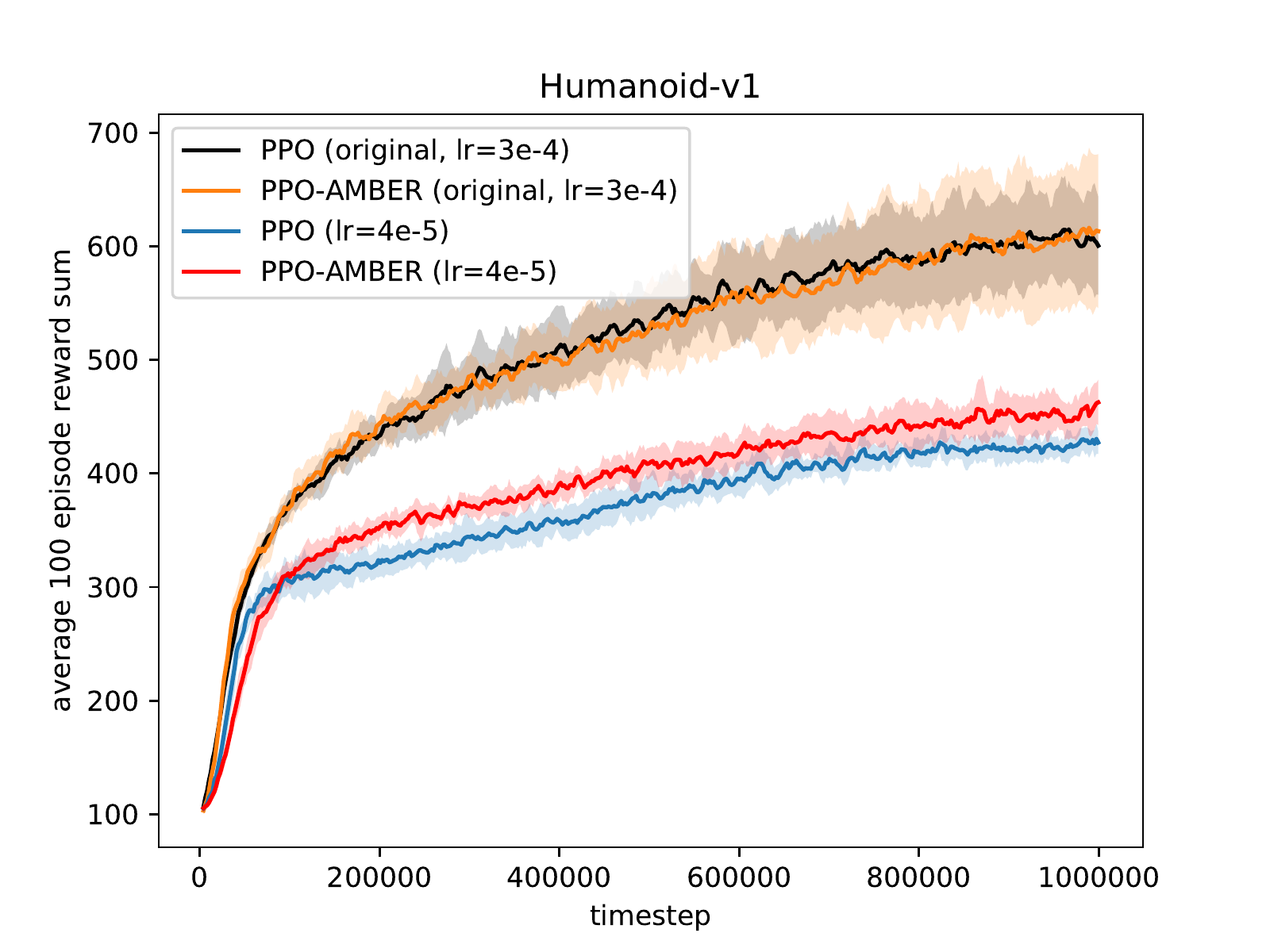}
	\includegraphics[width=0.4\textwidth]{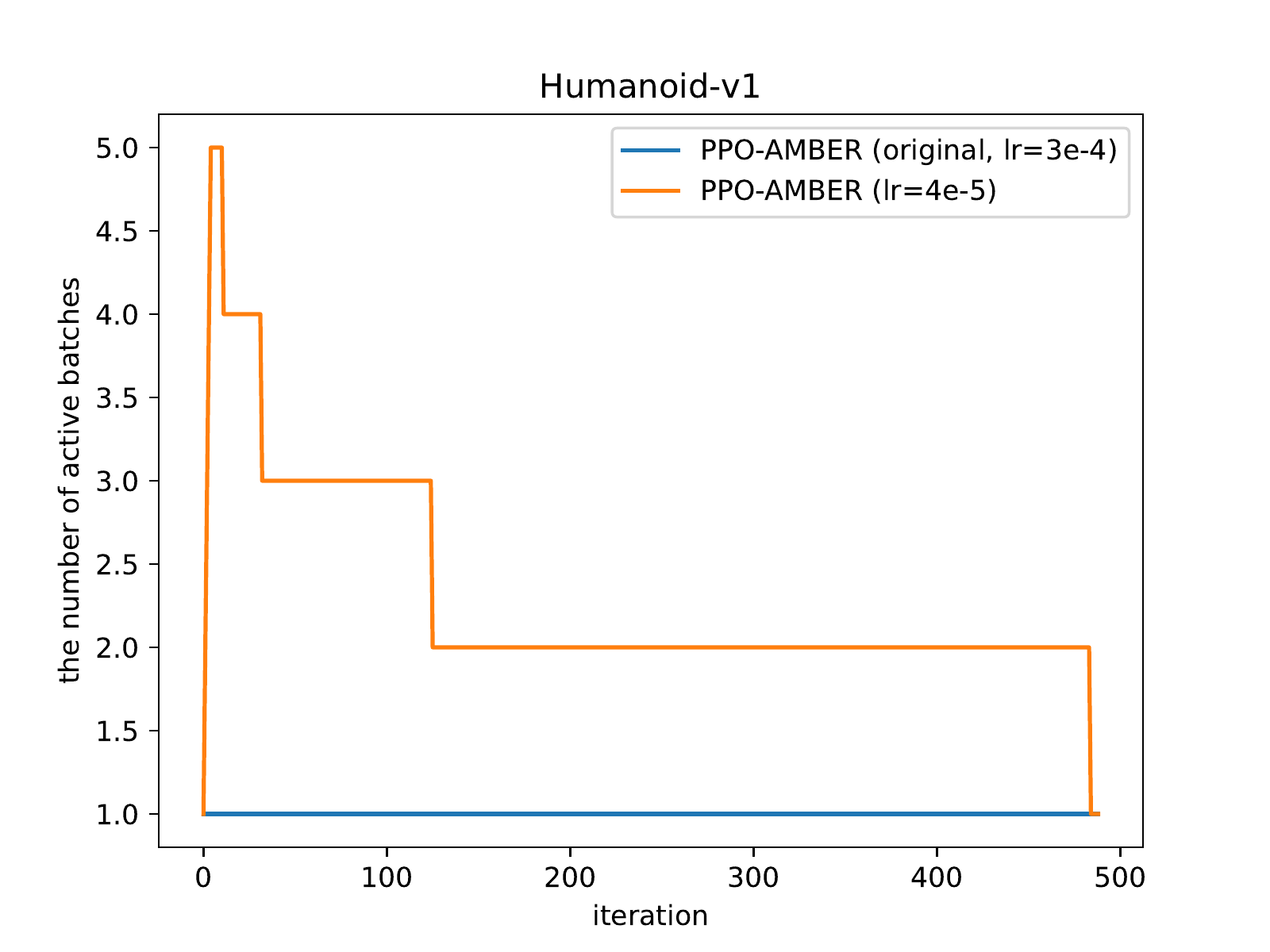}
	\caption{The evaluation of AMBER and the corresponding number of active batches}
\end{figure}

The different dimensions of the action spaces of different tasks much affect different IS weights for different tasks.  The IS weight can be factorized as $R_t(\tilde{\theta}) = \frac{\pi_{\tilde{\theta}}(a_t|s_t)}{\pi_{\theta}(a_t|s_t)}=\prod_{k=1}^K\frac{\pi_{\tilde{\theta}}(a_{t,k}|s_t)}{\pi_{\theta}(a_{t,k}|s_t)}$, where $K$ is the action dimension, $a_{t,k}$ is the $k$-th element of $a_t$, and $\pi_\theta(a_{t,k}|s_t)=\frac{1}{\sqrt{(2\pi)\sigma_k}}\exp(-(\mu(s_t;\phi)_k-a_{t,k})^2/2\sigma_k^2)$. Thus,  the IS weight increases as $K$ increases, if the change of $\mu_k$ and $\sigma_k$ is similar for each action dimension. Fig. \ref{fig:BatchAvgWeight} shows this behavior (the action dimension - Pendulum : 1, BipedalWalkerHardcore : 4, Humanoid : 17).
Hence,  the IS weight for Humanoid for old samples is too large and using the current sample only is best for Humanoid (and HumanoidStandup). With $\epsilon_b = 0.25$ across all tasks, samples with the IS weight larger than 1.25 are not used.
Note that small learning rates reduce the change of $\mu_k$ and $\sigma_k$, and consequently reduce the IS weight.
Hence, in order to apply AMBER to the harder tasks with high action dimensions,
 we consider reducing the learning rate from $(3\cdot10^{-4})$ to $(4\cdot10^{-5})$ and re-applying AMBER.
 The result is shown in Fig. \ref{fig:LR}.
  As expected, AMBER with the small learning rate uses a larger number of batches than the original AMBER due to the reduced IS weights, and AMBER improves the performance of PPO with the small learning rate. However, the performance behavior of PPO-AMBER is different for tasks. For Humanoid, reducing the learning rate harms the performance more than the improvement by AMBER. So, the overall performance with the reduced learning rate is worse. On the other hand, for HumanoidStandup, reducing the learning rate enhances the performance, and AMBER further improves the performance.  These results suggest that AMBER is efficient for small action-dimension tasks or sufficiently small learning rates so that IS weights are not too large.

\end{document}